\documentclass[runningheads]{llncs}

\usepackage{eccv}

\usepackage[british]{babel}

\usepackage{eccvabbrv}

\usepackage{graphicx}
\usepackage{booktabs}

\usepackage[accsupp]{axessibility}  

\usepackage{hyperref}

\usepackage{orcidlink}

\usepackage{mathtools} 
\usepackage{amssymb}
\usepackage{array, booktabs}

\usepackage{graphicx}
\usepackage[figuresright]{rotating}
\usepackage{times}
\usepackage{epsfig}
\usepackage{graphbox}
\usepackage{appendix} 
\usepackage{multirow}
\usepackage{subcaption}
\usepackage[super]{nth}
\usepackage{float}
\usepackage{pifont}

\usepackage[ruled, linesnumbered]{algorithm2e}

\usepackage[capitalize]{cleveref}
\crefname{section}{Sec.}{Secs.}
\Crefname{section}{Section}{Sections}
\Crefname{table}{Table}{Tables}
\crefname{table}{Tab.}{Tabs.}

\DeclareMathOperator*{\argmax}{argmax}
\DeclareMathOperator*{\soft}{softmax}

\DeclareMathOperator*{\concat}{concatenate}
\DeclareMathOperator*{\mean}{mean}

\begin{document}
\title{Uncertainty Estimation in Deep Learning for Panoptic Segmentation} 

\titlerunning{Uncertainty Estimation in D.L. for Panoptic Segmentation}

\author{Michael Smith\orcidlink{0009-0007-8611-9086} \and
Frank Ferrie}

\authorrunning{M.~Smith and F.~Ferrie}

\institute{Centre for Intelligent Machines\\
McGill University, Montreal QC, Canada\\
\email{michael.smith6@mail.mcgill.ca}\\
}

\maketitle

\begin{abstract}
As deep learning-based computer vision algorithms continue to advance the state of the art, their robustness to real-world data continues to be an issue, making it difficult to bring an algorithm from the lab to the real world. Ensemble-based uncertainty estimation approaches such as Monte Carlo Dropout have been successfully used in many applications in an attempt to address this robustness issue.  Unfortunately, it is not always clear if such ensemble-based approaches can be applied to a new problem domain.  This is the case with panoptic segmentation, where the structure of the problem and architectures designed to solve it means that unlike image classification or even semantic segmentation, the typical solution of using a mean across samples cannot be directly applied.  In this paper, we demonstrate how ensemble-based uncertainty estimation approaches such as Monte Carlo Dropout can be used in the panoptic segmentation domain with no changes to an existing network, providing both improved performance and more importantly a better measure of uncertainty for predictions made by the network.  Results are demonstrated quantitatively and qualitatively on the COCO, KITTI-STEP and VIPER datasets.
\end{abstract}

\section{Introduction}
Segmentation - or parsing the content of an image - has been a task in Computer Vision for many decades with a range of applications \cite{levine_scene_1973, marr_vision_1982, rosenfeld_chapter_1982, tighe_scene_2014}. \cite{kirillov_panoptic_2019} recently introduced Panoptic segmentation, a task that unifies semantic segmentation and instance segmentation (illustrated in \cref{app:sec:panoptic_viz} in the Supplementary Material) and can be viewed as a modern reformulation of older segmentation tasks, with the goal of generating a coherent scene interpretation remaining unchanged. More specifically, panoptic segmentation requires two labels consistent with one another to be assigned to each pixel in an image: a class label (from the semantic segmentation task) and an instance label (from the instance segmentation task). The class label specifies what the object is (car, sky, building), while the instance label differentiates between countable objects (such as cars). This consistency requirement differentiates panoptic segmentation from a multitask scenario with semantic and instance segmentation, where the label assignments may conflict with each other. As with other segmentation tasks, the main application of panoptic segmentation is to serve as input for a higher level process, such as autonomous navigation in a vehicle.

\begin{figure}
\centering
\begin{subfigure}{\textwidth}
    \centering
    \includegraphics[trim={150px 0 295px 60px}, clip, width=0.49\textwidth]{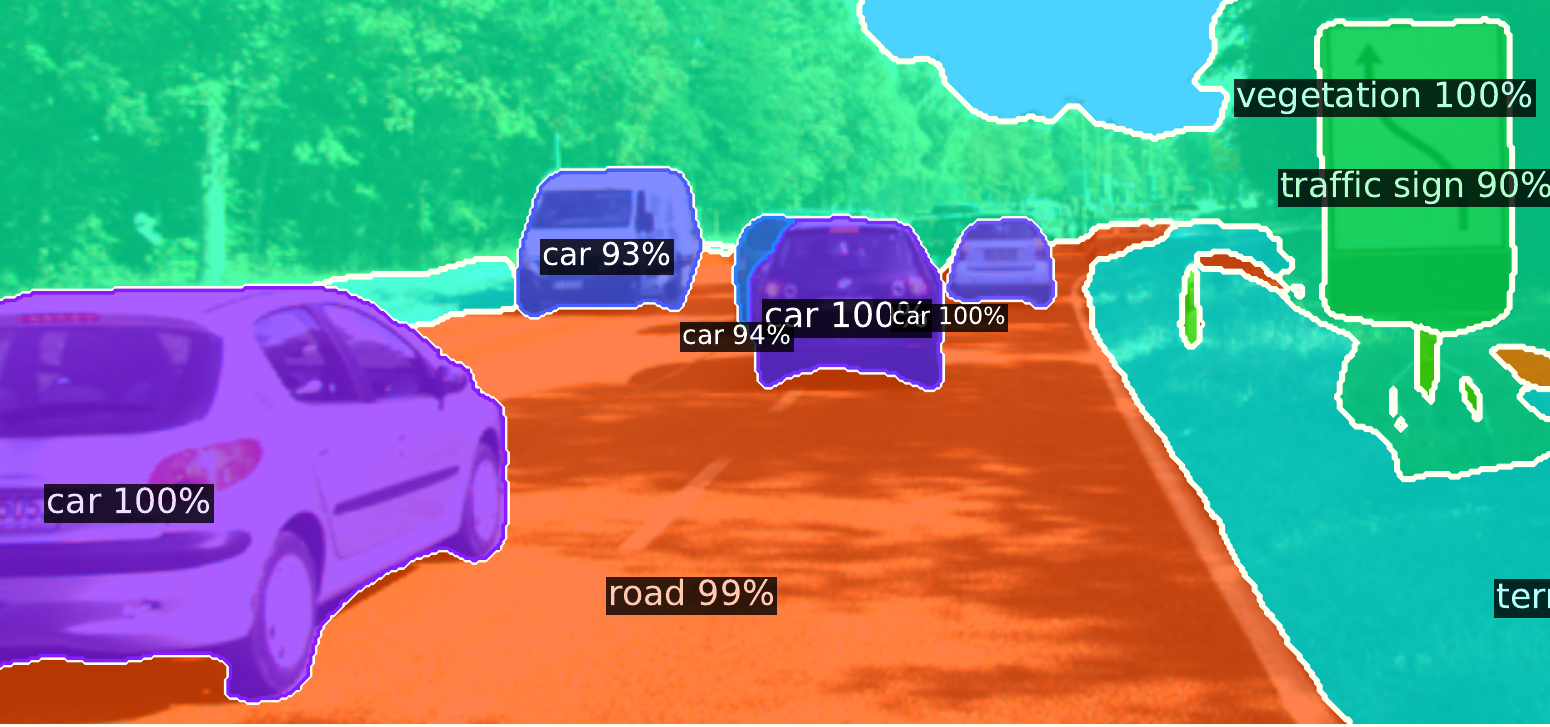}
    \includegraphics[trim={150px 0 295px 60px}, clip, width=0.49\textwidth]{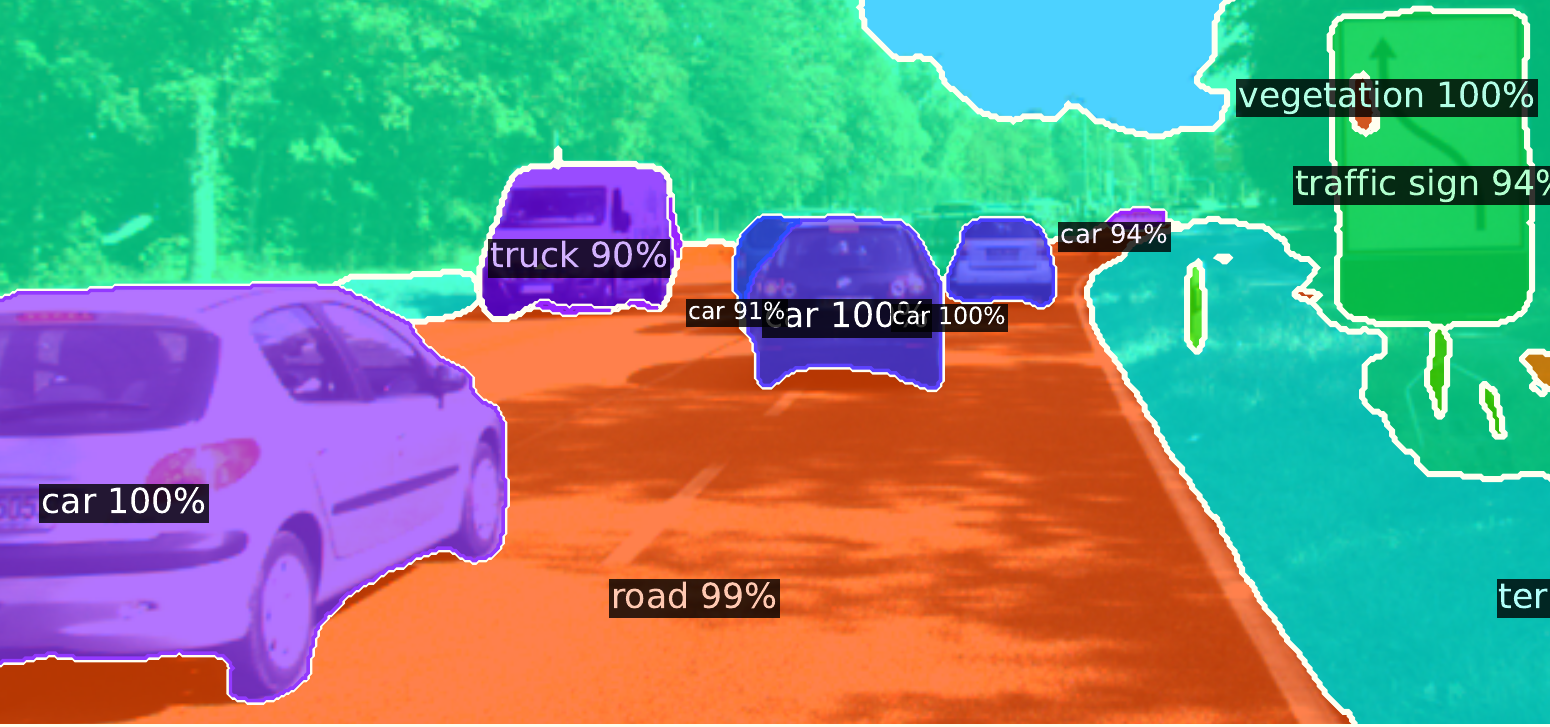}
    \caption{Baseline approach.}
    \label{fig:kitti_truck_car_motivation_base}
\end{subfigure}
\begin{subfigure}{\textwidth}
    \centering
    \includegraphics[trim={150px 0 295px 60px}, clip, width=0.49\textwidth]{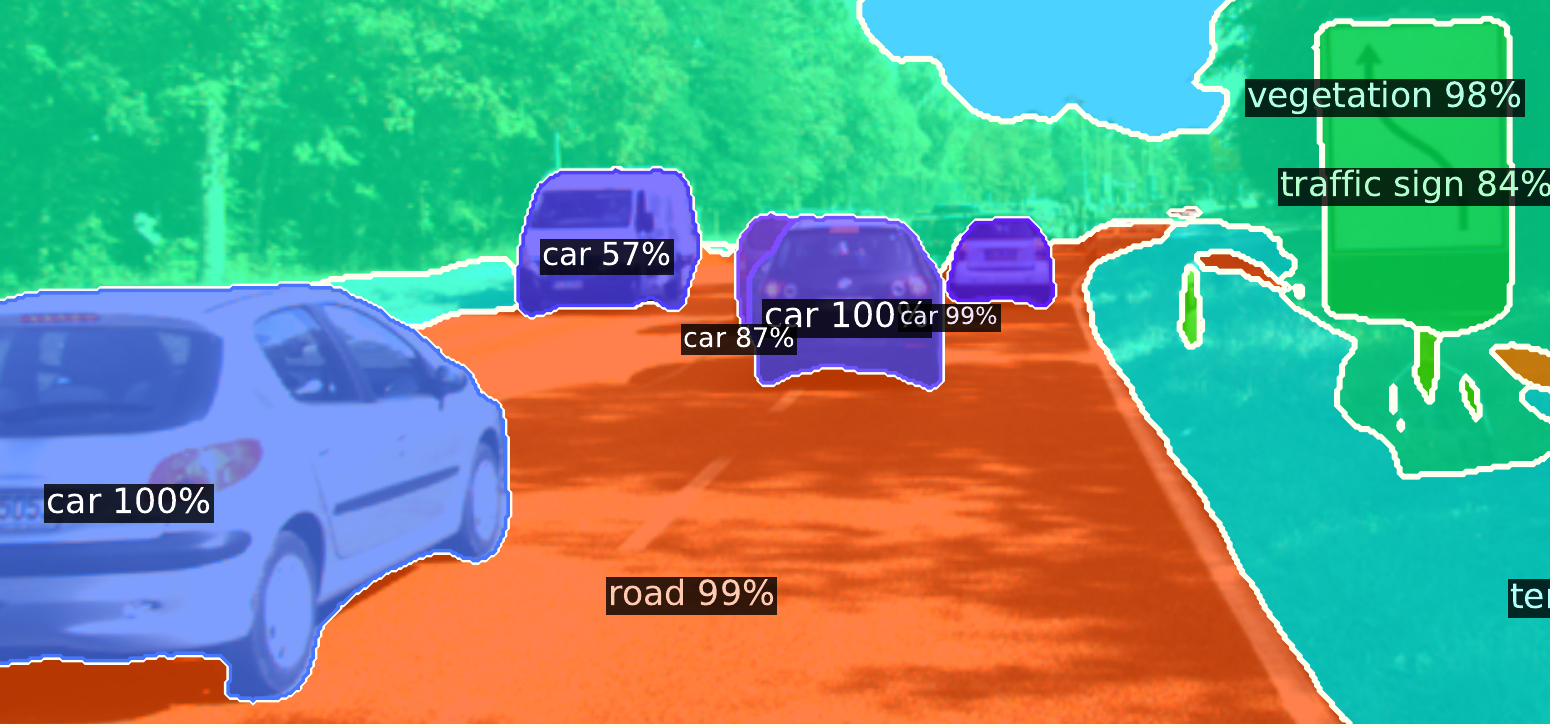}
    \includegraphics[trim={150px 0 295px 60px}, clip, width=0.49\textwidth]{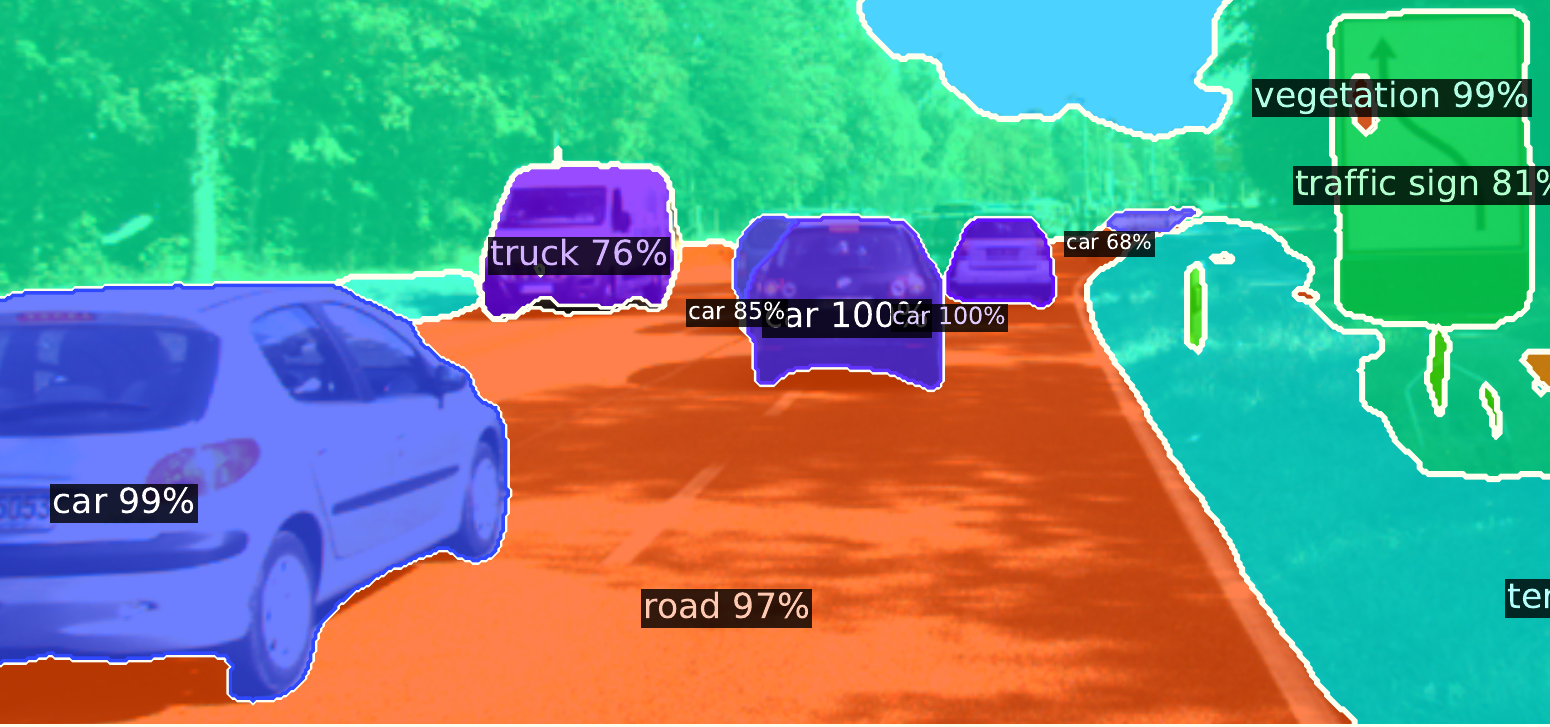}
    \caption{Our approach, with 15 Monte Carlo samples.}
    \label{fig:kitti_truck_car_motivation_ours}
\end{subfigure}
\caption{Two sequential frames (cropped for clarity) from the KITTI-STEP \cite{weber_step_2021} dataset. Note the change from identification of the oncoming vehicle from car to truck, and how our approach better captures the uncertainty in the prediction.}
\label{fig:kitti_truck_car_motivation}
\end{figure}

With all panoptic segmentation algorithms to our knowledge using neural networks (due to the complexity of the task and datasets), they unfortunately share a common issue, in that they can make mistakes in their predictions and they lack a reliable measure of confidence to qualify said predictions. We can see this in \cref{fig:kitti_truck_car_motivation_base}, where in the span of a single frame we see the prediction change from being very confident that the oncoming vehicle is a car to being very confident that it is a truck. This can be a problem that makes it difficult to rely on the predictions. Various attempts have been made to introduce such measures to neural networks in particular \cite{ghahramani_probabilistic_2015}, including ensembles, of which MC Dropout \cite{gal_dropout_2016} is but one example.

Typically, ensemble samples are generated from variations in the model parameters. Operators such as the mean can be used to generate a single prediction (averaging out noise) or Shannon entropy \cite{shannon_mathematical_1948} can be used to identify uncertainty in the predictions. Generally speaking, any disagreement between samples (or the lack thereof) can be a useful signal to a higher order process such as autonomous driving. A standard assumption that allows the aforementioned operators to be applied is that the samples are matched to one another, that is, they represent the same object. This is satisfied by default for some tasks, such as image classification, or semantic segmentation, where we have a per-pixel class map which provides an implicit correspondence \cite{huang_efficient_2018}. In our work, we investigate certain types of panoptic segmentation networks that generate a fixed set of predictions, where we would need to calculate the correspondence between each prediction belonging to each sample to determine if they represent the same object before we can apply operators such as the mean. A process for this does exist in other domains \cite{morrison_estimating_2019, weber_step_2021} using algorithms such as the Hungarian method \cite{kuhn_hungarian_1955}, but a straightforward implementation with respect to panoptic segmentation is not immediately evident.

As such, in this work we present a solution for solving the correspondence problem in an efficient and well-defined manner such that ensembles can be used for a selected set of approaches in the panoptic segmentation domain. This provides a more reliable measure of confidence, enabling the use of higher order processes that require such confidence measures. \cref{fig:kitti_truck_car_motivation_ours} provides a simple example of this, where our approach allows the accumulation of multiple samples (Monte Carlo samples in this case) which better capture the uncertainty in the predictions. Across multiple datasets, our approach typically provides a segmentation performance improvement compared to a baseline or other approaches. More importantly, it provides the ability to extract useful uncertainty information from multiple samples, via functions such as predictive entropy (\cref{eq:entropy_equation}). This enables downstream tasks such as autonomous driving to make better use of the panoptic segmentation results, taking into account the uncertainty as needed.

\section{Related Work}
\label{sec:related_work}
Panoptic segmentation \cite{kirillov_panoptic_2019} remains popular with recent datasets, challenges and frameworks emphasizing it \cite{lin_microsoft_2014, dosovitskiy_carla_2017, liao_kitti-360_2021}. As all pixels require a class label (\textit{e.g.} sky) but only some an instance ID (\textit{e.g.} car), a key differentiator between panoptic segmentation approaches is how these two label types are treated. We can broadly group them into three types: proposal-free, proposal-based and universal. Proposal-free (bottom-up) methods \cite{cheng_panoptic-deeplab_2020, wang_axial-deeplab_2020, bonde_towards_2021, li_fully_2021, qiao_detectors_2021} begin with semantic segmentation, followed by clustering of instance-related pixels. Proposal-based (top-down) methods \cite{kirillov_panoptic_fpn_2019, li_unifying_2020, mohan_efficientps_2021, yang_sognet_2020} on the other hand perform semantic and instance segmentation separately, and then fuse the (potentially conflicting) segments from each. Finally, a number of recent approaches use what we refer to as universal architectures \cite{cheng_masked-attention_2022, wang_max-deeplab_2021, cheng_per-pixel_2021, yu_cmt-deeplab_2022, li_panoptic_2022, zhang_k-net_2021} including DETR \cite{carion_end--end_2020}, which we build on for our experiments. They treat semantic and instance categories the same as much as possible, often via the use of a fixed set of category-agnostic predictions that are then processed similar to object detection \cite{zaidi_survey_2022}.

With respect to uncertainty estimation, ensemble-based (or sampling-based) methods such as MC Dropout \cite{gal_dropout_2016} and Deep Ensembles \cite{lakshminarayanan_simple_2017} remain the standard against which new approaches are evaluated despite their age \cite{malinin_shifts_2021, ashukha_pitfalls_2020, ovadia_can_2019}. Many approaches are presented as or can be viewed as ensembles \cite{yeo_robustness_2021, zhang_confidence_2019} and have seen wide ranging use \cite{ashukha_pitfalls_2020, kirsch_batchbald_2019} but are often considered computationally expensive due to the sample generation process \cite{mukhoti_deep_2022}. Other popular or recent approaches \cite{guo_calibration_2017,mukhoti_deep_2022,sensoy_evidential_2018} suffer from issues such as the inability to distinguish different sources of uncertainty or requiring non-trivial changes to the network architecture. 

When it comes to uncertainty estimation on segmentation tasks with modern neural networks, the vast majority of work \cite{huang_efficient_2018,mukhoti_deep_2021, zheng_rectifying_2021,nair_exploring_2020,kwon_uncertainty_2020} involves semantic segmentation or a close analog which avoids the complications inherent in panoptic segmentation. However, two recent approaches \cite{sirohi_uncertainty-aware_2023, deery_propandl_2023} do tackle the panoptic segmentation complications. Their approaches differ from ours in that they rely on deterministic approaches (as opposed to sampling) such as Evidential Deep Learning \cite{sensoy_evidential_2018}. As well, the network structures they use are substantially different from ours, and their solutions address issues specific to proposal-based architectures while ours focuses on universal architectures. Lastly, we note that the approach in \cite{morrison_estimating_2019} bears some similarities to ours, in that they make use of the Basic Sequential Algorithmic Scheme (BSAS) to match segments. However, they focus their work on instance segmentation rather than panoptic segmentation, of which instance segmentation is a subset. We also optimize BSAS for efficiency, whereas they appear to apply it directly. Furthermore, our results are much more comprehensive and coherent, clearly demonstrating the potential benefits sample-based uncertainty estimation methods can bring to panoptic segmentation. By contrast, they appear to mix the task of instance segmentation with metrics intended for object detection, and as such the benefits and the broader generalizability of their approach is unclear. Finally, to the best of our knowledge, all aforementioned methods that allow uncertainty estimates in panoptic segmentation propose some variation of semantic, instance or spatial uncertainty, which are more ad-hoc than the well-known uncertainty functions such as predictive entropy that we apply.

\section{Method}
\subsection{Problem}
\label{subsec:problem}
The primary question we need to answer is \textit{How can we obtain reliable measures of uncertainty in the panoptic segmentation domain?} We assume that ensemble models are used for uncertainty estimation (due to their popularity) and the panoptic segmentation architecture matches our definition of universal, rather than proposal-based or proposal-free. Such universal architectures generate a fixed number of predictions (each with their own softmax scores, masks and other attributes), which are then typically pruned to a small number of ``good'' ones. This network design style has also been used in other domains, such as object detection or instance segmentation. In the case of panoptic segmentation, it has the advantage of guaranteeing alignment between instance labels and semantic labels as they are generated \textbf{concurrently}. Concurrent assignment of the labels also allows them to inform each other as appropriate, potentially leading to superior results, and avoids forced error propagation. For example, if the class labels are wrong and frozen, subsequent assignment of instance labels may be difficult.

Given these constraints, when acquiring uncertainty estimates from a neural network using an ensemble with $Q$ models (samples), we have \cite{lakshminarayanan_simple_2017,gal_dropout_2016,ashukha_pitfalls_2020,malinin_uncertainty_2019_presentation,malinin2019uncertaintythesis}
\begin{equation}
    \left\{P(y \mid \boldsymbol{x}^*, \boldsymbol{\theta}^{(q)})\right\}^Q_{q=1}, \hspace{1em}\boldsymbol{\theta}^{(q)} \sim p(\boldsymbol{\theta}|\mathcal{D}).
    \label{eq:main_ensemble}
\end{equation}
With that, we need to obtain a single prediction that takes into account information from all available samples. Typically, this is done by obtaining the predictive posterior as the mean across samples \textit{i.e.}
\begin{equation}
    \label{eq:predictive_mean}
    P(y \mid \boldsymbol{x}^*, \mathcal{D}) = \frac{1}{Q} \sum^Q_{q=1} P(y \mid \boldsymbol{x}^*, \boldsymbol{\theta}^{(q)}).
\end{equation}
The uncertainty of this prediction can be obtained in several ways via statistics from the ensemble; for example, we can calculate the total uncertainty as the entropy \cite{shannon_mathematical_1948} of the predictive posterior:
\begin{equation}
\label{eq:entropy_equation}
\begin{split}
        \mathcal{H}\left[P(y \mid \boldsymbol{x}^*, \mathcal{D})\right] \approx \mathcal{H}\left[\frac{1}{Q} \sum^Q_{q=1} P(y \mid \boldsymbol{x}^*, \boldsymbol{\theta}^{(q)})\right], \\
        \boldsymbol{\theta}^{(q)} \sim p(\boldsymbol{\theta} \mid \mathcal{D}).
\end{split}
\end{equation}
Other metrics such as mutual information or sample entropy can also be calculated, each of which captures to varying degrees epistemic (model) or aleatoric (data) uncertainty. The former covers the possibility of multiple models being equally good fits to the data, while the latter covers labeling or measurement noise, where for example human annotators may disagree on the label for an image. Different tasks may require epistemic and/or aleatoric uncertainty \textit{e.g.} identifying out-of-distribution examples requires epistemic uncertainty \cite{kendall_what_2017}.

Critically, \cref{eq:predictive_mean} includes an implicit assumption that the prediction from each ensemble member can be matched with that of all other members.  In the case of image classification or semantic segmentation, this assumption holds as the frame of reference (image or pixel) is the same for all ensemble members, allowing \cref{eq:predictive_mean} to apply at that level. In the case of panoptic segmentation (and instance segmentation), this is not necessarily the case.

Consider that the output $O$ of the DETR \cite{carion_end--end_2020} neural network in its panoptic configuration can be expressed as
\begin{equation}
    O = \{L_{N \times C}, M_{N \times h \times w}\},
\end{equation}
where $L$ is a set of logit vectors on which the softmax function can be applied to obtain a class representation of each proposal $N \in N$, while $M$ is a mask associated to each proposal $n \in N$. $C$ represents the number of classes in the dataset. Each proposal $n$ represents a potential object in the image, such as a vehicle, building or a more general category such as the sky.  Being that $N$ is fixed, at inference time post-processing is required to preserve only the useful proposals and thus create a high quality panoptic segmentation (see \cref{app:subsec:baseline_impl_details} for details). When we use an ensemble, we now have multiple samples of the neural network output $\{O_1, \dots, O_N\}$, but no clear way to aggregate the samples, as each proposal $n$ may represent a different object in each sample and we do not know the correspondence between the instances.

\subsection{Our approach}
\label{sec:our_approach}
\begin{figure}
    \centering
    \includegraphics[width=\textwidth]{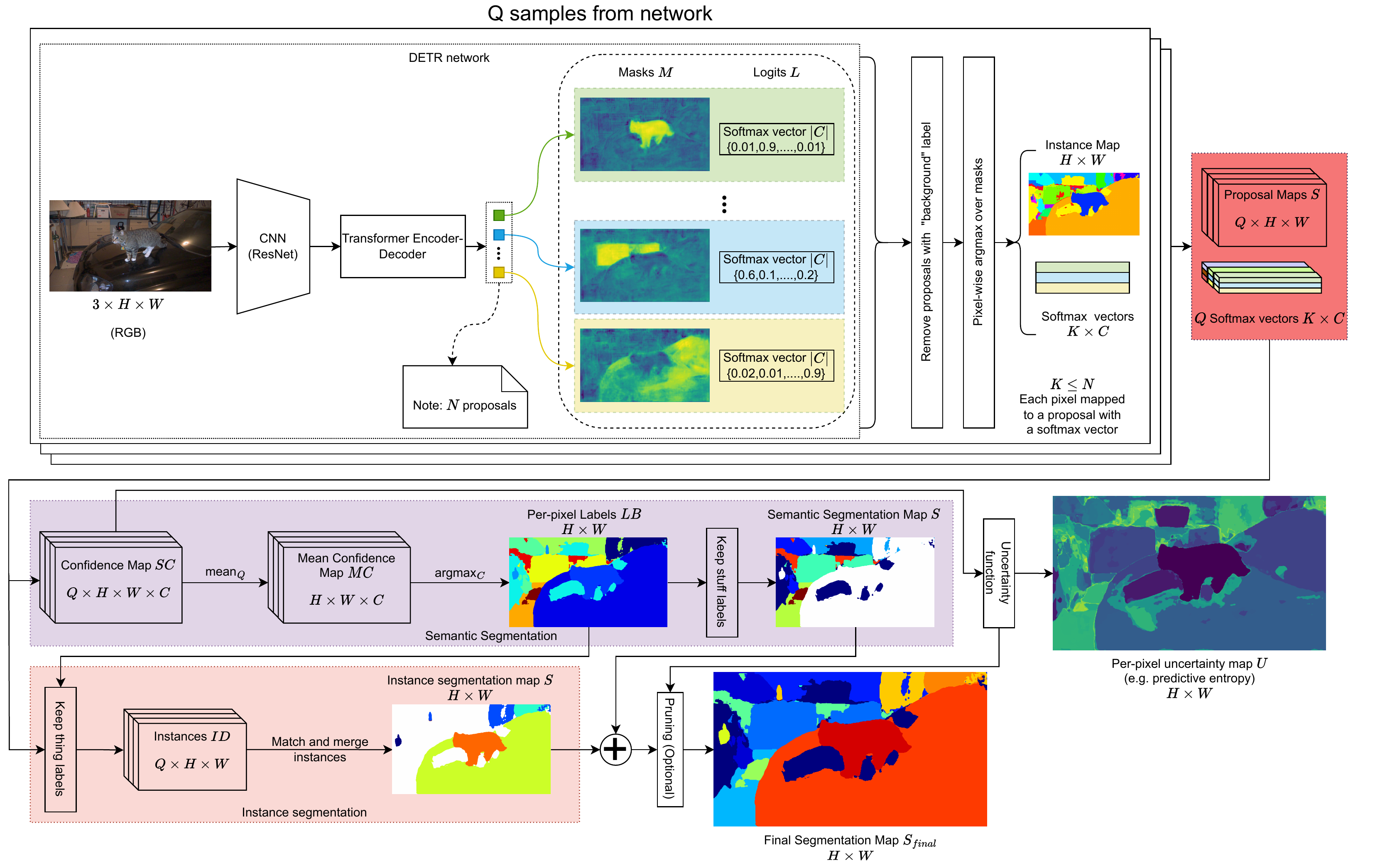}
    \caption{Overview of our approach. See text for details. Note that for conciseness and clarity, some implementation details have been simplified.}
    \label{fig:our_approach_diagram}
\end{figure}

Our task is to aggregate multiple samples collected from the neural network in such a way that we obtain both an acceptable panoptic segmentation result and a corresponding pixel-level uncertainty estimate, shown in \cref{fig:our_approach_diagram}.

The first step is to collect and process the samples from the network using MC Dropout \cite{gal_dropout_2016}, shown in the upper half of \cref{fig:our_approach_diagram}. We then perform some minimal processing adapted from the baseline approach to reduce the $N$ proposals to $K \le N$ as many proposals are essentially noise \cite{carion_end--end_2020}. This is shown in \cref{app:alg:perSampleSegmentation} (\texttt{perSampleSeg}) in the Supplementary Material, with the result being a set of softmax vectors and proposal maps, which link each pixel to an instance ID and a class label.

The second step involves semantic (stuff) segmentation, shaded in light purple in \cref{fig:our_approach_diagram} and shown in \cref{alg:stuffSegmentation}. While panoptic segmentation is a joint problem where the simultaneous assignment of semantic and instance labels is helpful, we find that decomposing the problem after acquiring our samples from the network is useful for both performance and speed. As such, we first deal with the simpler case of the semantic labels by applying the mean across samples to identify the class label at each pixel. If the pixel maps to a semantic class (no instance ID required), we are done as we only need the class label. If the pixel maps to a class where instance IDs are required, we proceed to the instance segmentation step.

The instance segmentation step is shaded in light pink in \cref{fig:our_approach_diagram} and shown in \cref{alg:thingSegmentation}. Our goal is to assign instance IDs to each pixel without discarding relevant information or being excessively computationally complex. Our process is a modified form of the Basic Sequential Algorithmic Scheme \cite{theodoridis_chapter_2009}: examining each instance in each sample, we compute the Intersection over Union score in image space between that instance and a list of instances. If the score is above some threshold, we consider it a match and merge the two instances. If we cannot find a match in the list, we consider it a new unseen instance and add it to the list.  When this process is complete, we are left with a list of instances, most of which will have been matched with others across samples. From this list, we can generate an instance segmentation map $S$ by assigning pixels to instances (thus matching them with a class and instance ID) in the order of the number of matches. 

The final step is to undo our prior decomposition and combine the output of the semantic and instance segmentation steps to create a complete segmentation map $S_{final}$, with all pixels being assigned a class label and some being assigned an instance ID where appropriate. This is shown in the bottom right of \cref{fig:our_approach_diagram} and in \cref{alg:generateSegmentation}. As we inherit the base network's tendency to generate a large number of false positives, we optionally prune the segmentation via thresholds on the confidence. Alternatively, we can apply more mathematically grounded measures of uncertainty (\textit{e.g.} predictive entropy) on the per-pixel uncertainty map and take appropriate action (\textit{e.g.} defer to a human) in areas where the model is uncertain.

\begin{algorithm}
\DontPrintSemicolon
\caption{generateSegmentation(Net)}
\label{alg:generateSegmentation}
\KwIn{Panoptic Neural network \textit{Net}}
\KwOut{Segment Map $S^{H \times W}$, Per-pixel uncertainty map $U^{H \times W}$}
\SetKwFunction{stuffseg}{stuffSeg}
\SetKwFunction{thingseg}{thingSeg}
\SetKwFunction{perSam}{perSampleSeg}
\SetKwFunction{thres}{prune}
\SetKwFunction{uncert}{uncertainty}
Gather $Q$ samples from network \textit{Net} resulting in:\;
Logits $L^{Q \times N \times C}$ and Masks $M^{Q \times N \times h \times w}$\;
\tcp{$C$: classes}
\tcp{$h$ \& $w$: input image dimensions}
\tcp{$N$: \# of proposals}
$S_{perSample}, I \leftarrow$ \perSam{$L$, $M$}\;
$S_{initial}, SC, MC \leftarrow $\stuffseg{$S_{perSample}$, $I$}\;
$S_{final} \leftarrow $\thingseg{$S_{perSample}$, $I$, $S_{initial}$, $MC$}\;
\tcp{Pruning optional}
$S_{final} \leftarrow $\thres{$S_{final}, SC$}\; \label{alg:line:pruning}
\tcp{Multiple uncertainty functions possible}
$U \leftarrow $ \uncert{$SC$}\;
\KwRet{$S_{final}$, $U$}
\end{algorithm}

\begin{algorithm}
\DontPrintSemicolon
\caption{stuffSeg(\textit{S}, \textit{I})}
\label{alg:stuffSegmentation}
\KwIn{Per-sample Proposal Map $S^{Q \times H \times W}$, List length $Q$, of proposal softmax vectors $I^{K \times C}$}
\KwOut{Semantic Segmentation Map $S^{H \times W}$, Confidence map $SC^{Q \times H \times W \times C}$, Mean confidence map $MC^{H \times W \times C}$,}
\tcp{$SC$: Eq. (1)}
$SC^{Q \times H \times W \times C} \leftarrow \mathbf{0}$\;
\ForEach{Sample $q \in Q$}{
    \ForEach{Proposal $k \in K$}{
        $SC_{q, i, j} \leftarrow I_{q, k} \,\, \forall \,\, i,j$ where $S = k$\;    
    }
}
\tcp{$MC$: Eq. (2)}
$MC^{H \times W \times C} \leftarrow \mean_Q{SC}$\; 
\tcp{$LB$: per-pixel labels}
$LB^{H \times W} \leftarrow \argmax_C{MC}$\; 
$S^{H \times W} \leftarrow \mathbf{0}$\;
\ForEach{label $lb \in LB$}{
    \If{$lb \subseteq $ stuff}{
        $S_{i,j} \leftarrow lb \,\, \forall \,\, i,j$ where $lb = LB_{i,j}$\;
    }
}
\KwRet{$S$, $SC$, $MC$}
\end{algorithm}

\begin{algorithm}
\DontPrintSemicolon
\SetKwFunction{sort}{sortByMergers}
\caption{thingSeg(\textit{S}, \textit{I}, \textit{S\textsubscript{initial}}, MC)}
\label{alg:thingSegmentation}
\KwIn{Per-sample Proposal Map $S^{Q \times H \times W}$, List length $Q$, of proposal softmax vectors $I^{K \times C}$,
Initial Semantic Segmentation Map $S^{H \times W}_{initial}$, Mean Confidence Map $MC^{H \times W \times C}$}
\KwOut{Complete Segmentation Map $S^{H \times W}$}
$LB^{H \times W} \leftarrow \argmax_C{MC}$\; 
$ID \leftarrow \mathbf{0}$\;
$ID^{Q \times H \times W}_{i,j} \leftarrow S_{i,j} \,\, \forall \,\, i,j$ where $LB_{i,j} \subseteq$ things\;
\tcp{$SEG$: list of segments}
$SEG \leftarrow \mathbf{0}$\;

\tcp{Below similar to BSAS}
\ForEach{Sample $q \in Q$}{
    \ForEach{Proposed instance $g \in I_q$}{
        Match $\leftarrow$ False\;
        \ForEach{Instance $h \in SEG$}{
            \If{IoU $\geq$ threshold}{
                Match $\leftarrow$ True\;   
                Merge $g$ into $h$\;
                Track \# of merges\;
            }
        }
        \If{Match $=$ False}{
            Add $g$ to $SEG$\;    
        }
    }
}
$SEG \leftarrow$ \sort{$SEG$}\;
$S \leftarrow S_{initial} \,\, \forall \,\, i,j$ where $S_{initial} \neq 0$, else $\mathbf{0}$\;
\ForEach{Instance $h \in SEG$}{
    \If{Merge count of $h \geq $ threshold }{
        $S \leftarrow S \cup ((S = 0) \cap h)$
    }
}
\KwRet{$S$}
\end{algorithm}

\section{Experimental Results}
\subsection{General implementation}
\label{sec:general_implementation}
Our framework uses the DETR model \cite{carion_end--end_2020} with the ResNet50 backbone. We use the COCO \cite{lin_microsoft_2014}, KITTI-STEP \cite{weber_step_2021} and VIPER \cite{richter_playing_2017} datasets. For the COCO dataset, a pre-trained model from \cite{carion_end--end_2020} is used. For other datasets, we train a custom model. Using the DETR approach in its original configuration represents the baseline. We further adapt and implement a particular variant of the Hungarian algorithm known as the Jonker–Volgenant algorithm to serve as an additional baseline \cite{jonker_shortest_1987}. All datasets are fairly complex and varied, containing thousands of images. We also create ``corrupted'' versions of all datasets following ImageNet-C \cite{hendrycks_benchmarking_2019}, except we only use one corruption where we combine the Gaussian and Shot Noise corruptions into one. Three versions of the dataset are generated with said corruption type applied at severity levels 1, 2 and 3. Please see \cref{app:sec:extra_implementation_details,app:sec:additional_results} in the Supplementary Material for more implementation details, including the use of a variant of the baseline without pruning and additional results from some experiments. As well, note that code will be published on acceptance.

For evaluation, we use the Panoptic Quality (PQ) metric as proposed in \cite{kirillov_panoptic_2019} and typically used to evaluate panoptic segmentation performance in the literature \cite{chen_generalist_2023,xu_open-vocabulary_2023}. It is defined as:
\begin{equation}
    \label{eq:pqcat}
    \text{PQ}_{\text{class}} = \frac{\sum_{(p,g) \in \text{TP}} \text{IoU}(p,g)}{|\text{TP}| + \frac{1}{2}|\text{FP}| + \frac{1}{2} |\text{FN}|}.
\end{equation}
In this metric, segment matching is done between ground truth annotations and predictions, with a true positive considered if the IoU $\geq 0.5$. True positives, false positives and false negatives are calculated on a per class basis following \cref{eq:pqcat}. We present our results as a mean across all classes. True negatives are not defined in this context. The PQ metric can also be decomposed into segmentation quality (SQ) and recognition quality (RQ) by
\begin{equation}
        \text{PQ} = \underbrace{\frac{\sum_{(p,g) \in \text{TP}} \text{IoU}(p,g)}{|\text{TP}|}}_{\text{SQ}} \times \underbrace{\frac{|\text{TP}|}{|\text{TP}| + \frac{1}{2}|\text{FP}| + \frac{1}{2} |\text{FN}|}}_{\text{RQ}}.
\end{equation}

\subsection{Basic Performance}
\label{sec:performance}

\begin{table}
    \centering
    \begin{tabular}{lll|lll|lll}
        \toprule
        & & & \multicolumn{3}{c}{Clean} & \multicolumn{3}{c}{Corrupted}\\
        \midrule
        & App. & Samp. & PQ$\uparrow$ & SQ$\uparrow$ & RQ$\uparrow$ & PQ$\uparrow$ & SQ$\uparrow$ & RQ$\uparrow$\\
        \midrule
        \multirow{5}{*}{\rotatebox{90}{COCO}} & Base & N/A & 43.4 & 79.3 & 53.8 & 28.9 & 75.4 & 36.8\\
        & Hng. & 5 & 39.5±0.1 & 78.8±0.1 & 49.0±0.1 & 28.1±0.0 & 76.2±0.2 & 35.7±0.1\\
        & Hng. & 15 & 40.7±0.1 & 79.1±0.1 & 50.4±0.1 & 28.9±0.1 & 76.3±0.2 & 36.7±0.1\\
        & Ours & 5 & 44.1±0.1 & 79.5±0.1 & \textbf{54.5±0.1} & \textbf{30.6±0.1} & 76.8±0.2 & \textbf{38.7±0.1}\\
        & Ours & 15 & \textbf{44.4±0.1} & \textbf{79.9±0.1} & \textbf{54.5±0.1}& \textbf{30.5±0.1} & \textbf{77.1±0.1} & 38.4±0.1\\
        \midrule
        \multirow{5}{*}{\rotatebox{90}{KITTI}} & Base & N/A & 39.9 & \textbf{77.5} & 50.2 & 19.3 & 54.9 & 26.0\\
        & Hng. & 5 & 36.1±0.2 & 76.5±0.2 & 45.4±0.2 & 17.1±0.1 & \textbf{58.3±1.2} & 23.1±0.1\\
        & Hng. & 15 & 37.6±0.1 & 76.9±0.1 & 47.2±0.1 & 17.8±0.1 & \textbf{58.1±0.2} & 24.1±0.1\\
        & Ours & 5 & \textbf{41.8±0.1} & 76.9±0.1 & \textbf{52.7±0.2} & 20.0±0.0 & 57.4±1.1 & \textbf{27.1±0.0}\\
        & Ours & 15 & \textbf{41.9±0.1} & 77.0±0.1 & \textbf{52.7±0.1} & \textbf{20.1±0.0} & 57.7±0.9 & \textbf{27.1±0.0}\\
        \midrule
        \multirow{5}{*}{\rotatebox{90}{VIPER}}
        & Base & N/A & 31.1 & 74.0 & \textbf{39.7} & \textbf{10.3} & 66.8 & \textbf{14.5}\\
        & Hng. & 5 & 27.0±0.0 & 73.1±0.1 & 34.6±0.0 & 8.4±0.0 & 68.8±0.2 & 12.1±0.0\\
        & Hng. & 15 & 28.4±0.0 & 73.6±0.1 & 36.3±0.0 & 9.2±0.0 & 69.2±0.2 & 13.1±0.0\\
        & Ours & 5 & 31.1±0.0 & 74.2±0.1 & 39.5±0.0 & 10.2±0.0 & \textbf{69.5±0.1} & 14.3±0.0\\
        & Ours & 15& \textbf{31.3±0.0} & \textbf{74.6±0.0} & 39.6±0.0 & 10.1±0.0 & \textbf{69.7±0.2} & 14.2±0.0\\
        \bottomrule
    \end{tabular}
    \caption{Basic performance. Results are presented as $\mu\,$±$\,\sigma$ over 10 runs when stochastic. \textit{Corrupted} columns represent the use of Gaussian and shot noise (severity 1). \textit{Hng.} refers to the Hungarian method.}
    \label{tab:eval_table}
\end{table}

For our first experiment, shown in \cref{tab:eval_table}, we evaluate our approach on the KITTI-STEP, COCO and VIPER datasets.

The key findings are that first, adding Gaussian and Shot noise clearly decimates performance across the board regardless of approach, and gathering additional MC Dropout samples from the network is not able to compensate for poor network performance from the added noise. Second, our approach generally leads to a slight performance increase over baseline (with one exception) and a much more significant increase over the Hungarian algorithm, on both clean and noisy data. Matching or exceeding the baseline is welcome, as we only use the uncertainty estimates we gain in later experiments. This demonstrates our approach's ability to correctly integrate information from multiple samples. With respect to the Hungarian algorithm's performance, please see \cref{app:sec:hungarian_performance}. In general, we note that performance drops when using multiple samples have been shown to occur in other work \cite{deery_propandl_2023}; a situation which our approach overall manages to avoid. Third, we see that while adding more samples does result in a slight performance increase, $5$ samples is typically sufficient to generate a reasonable segmentation result in this context, with the additional samples adding computational complexity for limited benefit.

\subsection{Application of uncertainty estimates}
\label{sec:applied_uncertainty_estimate}
\begin{sidewaysfigure}
    \centering
    \includegraphics[width=0.80\textwidth]{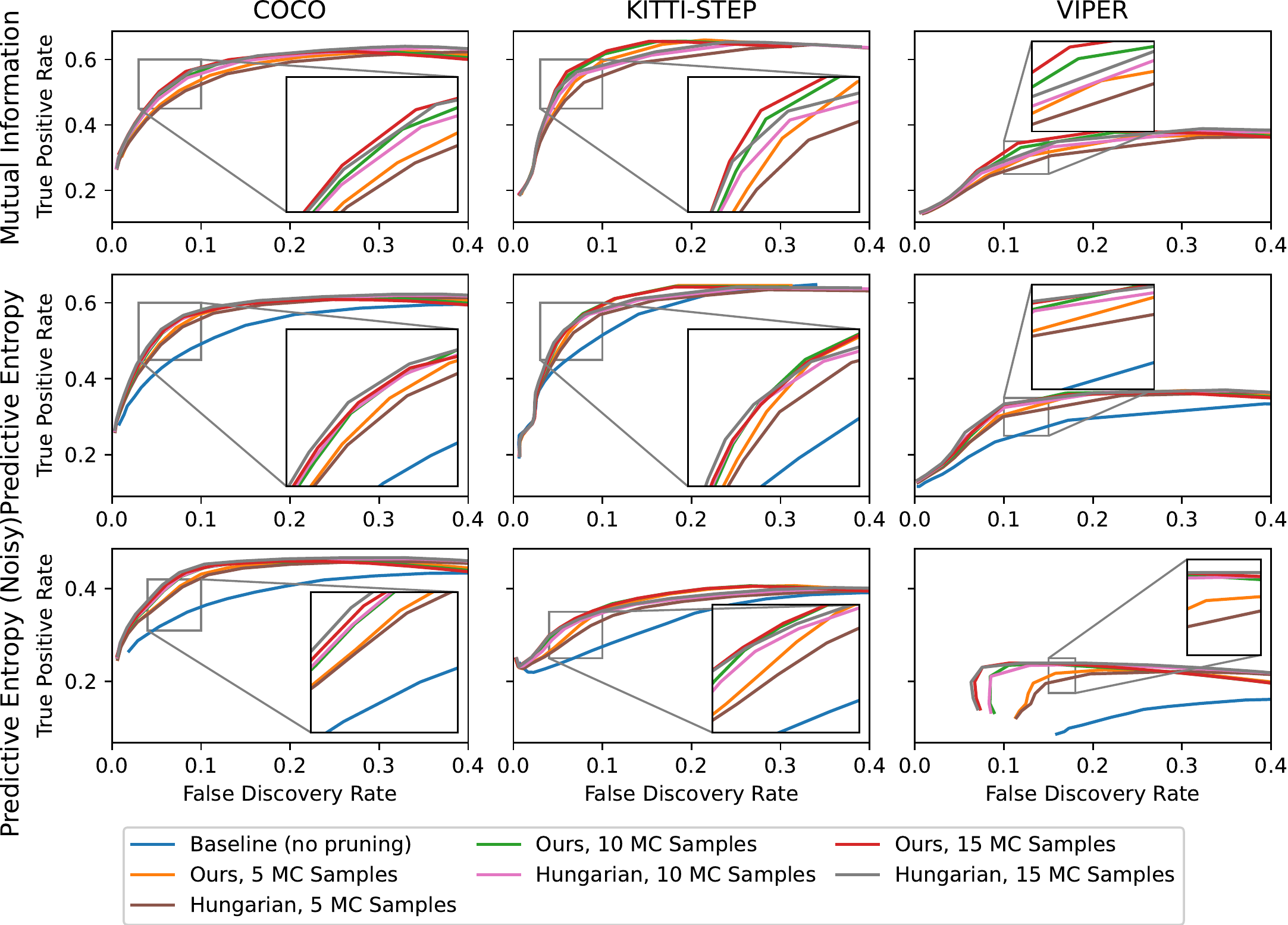}
    \caption{Using various uncertainty measures to identify and remove the most uncertain pixels in the dataset, we plot the True Positive Rate versus the False Detection Rate. Better uncertainty estimates result in a greater area under the curve. Results are collected over 10 runs and averaged; standard deviation is visually negligible. \textit{Noisy} indicates that Gaussian and Shot Noise at severity level 1 is applied to the images provided to the network. The baseline is not shown in the mutual information plots as the measure requires more than 1 sample to calculate.}
    \label{fig:graphs_uncertainty_main}
\end{sidewaysfigure}

In \cref{fig:graphs_uncertainty_main}, for all three datasets we plot ROC-like (Receiver Operating Characteristic) curves for an experiment where we vary a threshold that determines how uncertain a pixel must be before it is removed (ignored during evaluation as if no ground truth annotation existed) from the dataset. We remove up to 60-95\% of the pixels in the validation set which are marked as uncertain, depending on the dataset and/or use of noisy data; removing any more destabilizes the evaluation metric. A greater area under the curve is desirable. We plot the predictive entropy and mutual information measures with clean data, and predictive entropy for Gaussian and Shot Noise at severity level 1. There are four observations we can make. The first is that as expected, our approach using predictive entropy from Monte Carlo samples (\cref{eq:entropy_equation}) reliably outperforms the baseline using softmax entropy \cite{mukhoti_deterministic_2021}. Our approach does perform on par or better than the Hungarian algorithm, albeit with some exceptions in a few cases. We also note that performance improvements with MC Dropout drop off or even reverse slightly towards the right of the plots; this is expected as it is in line with other similar work \cite{nair_exploring_2020}. Furthermore, there is a mismatch between how we remove uncertain data at the \textit{pixel} level but our evaluation relies on \textit{segment} matching. The second observation is that adding more samples provides a better uncertainty estimate, as expected, but with diminishing returns as more samples ($\geq10$) are used. The third observation is that in line with \cref{tab:eval_table}, adding noise generally decimates performance and ensemble uncertainty estimation methods that rely on MC Dropout are not necessarily any more resilient to it. The final observation is that both predictive entropy and mutual information, which capture epistemic uncertainty, perform well. This demonstrates that epistemic uncertainty is being properly modelled, as expected given MC Dropout's focus on it.

\begin{figure}
\centering
\begin{subfigure}{0.43\textwidth}
    \includegraphics[width=\textwidth]{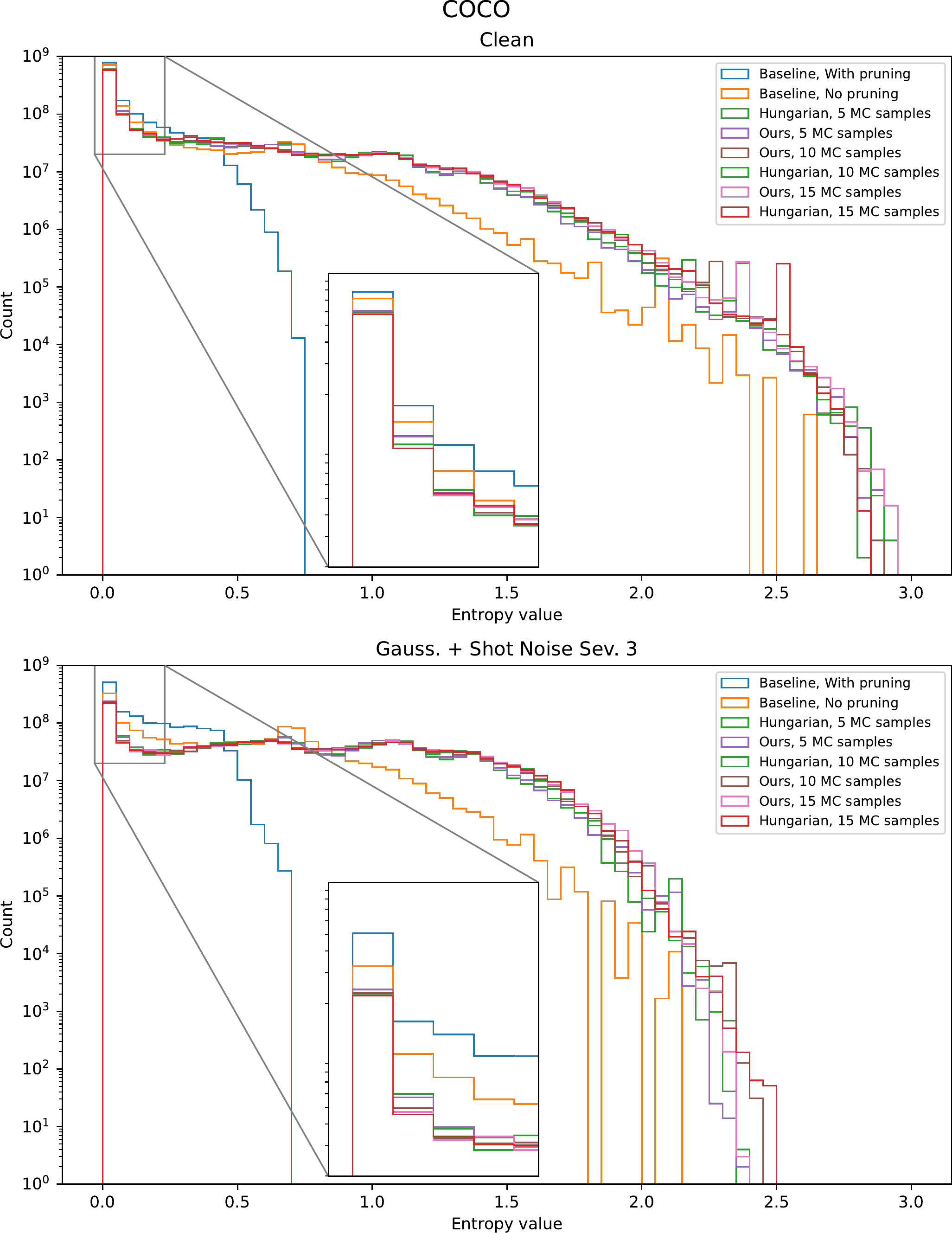}
    \label{fig:entropy_COCO}
\end{subfigure}
\begin{subfigure}{0.43\textwidth}
    \includegraphics[width=\textwidth]{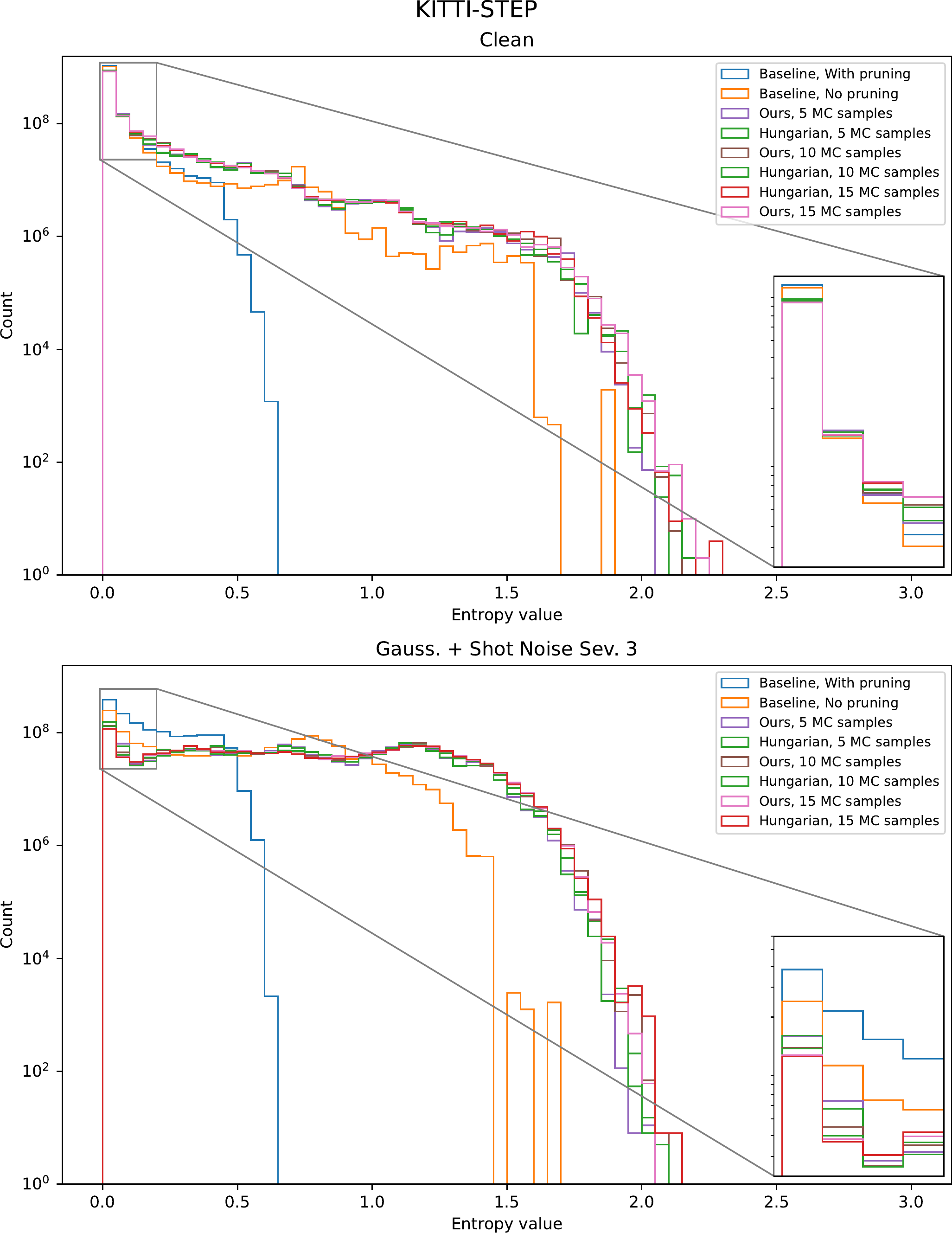}
    \label{fig:entropy_KITTI}
\end{subfigure}
\caption{Binning and plotting predictive entropy (ours) or softmax entropy (baseline) for our approach with Monte Carlo dropout and the baseline respectively. Top: clean data. Bottom: Gaussian and Shot noise at high severity. Note that the y-axis follows log scale.}
    \label{fig:pred_entropy_hist}
\end{figure}

In \cref{fig:pred_entropy_hist}, following a similar procedure as \cite{lakshminarayanan_simple_2017} we bin and plot the predictive entropy values (for the baseline, softmax entropy \cite{mukhoti_deterministic_2021}) generated by predictions for all pixels from all images in the COCO and KITTI-STEP datasets. We do this for both clean data and with a large amount of Gaussian and Shot noise added. We plot the baseline following \cite{carion_end--end_2020} as well as with pruning disabled. With a better measure of uncertainty, fewer predictions should be overconfident and there should be a high degree of uncertainty under abnormal conditions such as noise, where we expect the network to perform worse. In particular, under noisy conditions we would expect to see a more uniform distribution. Overall, the results on both datasets validate this hypothesis. First, the default baseline applies thresholds that result in only the more confident predictions remaining, thus decimating the available uncertainty information. When we disable those thresholds, we see that under normal conditions with clean data, both our approach and the Hungarian method result in a more uniform spread of values compared to that variant of the baseline. Under noisy conditions, this effect is further magnified, indicating a better representation of the (unknown) true conditional probabilities which should be more uniform. Interestingly, for the COCO dataset only we see fewer very uncertain predictions in the $2.5-3$ range when noise is added, but this occurs uniformly regardless of approach and is thus likely caused by the base network model. We also note that while our approach can work with any ensemble, our evaluation with MC Dropout will naturally inherit any limitations as shown by \cite{lakshminarayanan_simple_2017}.

\subsection{Qualitative Results}
\label{sec:qualitative_results}
\begin{figure}
\def\qual_width{0.245\textwidth}
\centering
\begin{subfigure}[b]{\qual_width}
    \includegraphics[width=\textwidth]{Figures/NoUncertainty/COCO/RGB/000000168330.jpg}
\end{subfigure}
\begin{subfigure}[b]{\qual_width}
    \includegraphics[width=\textwidth]{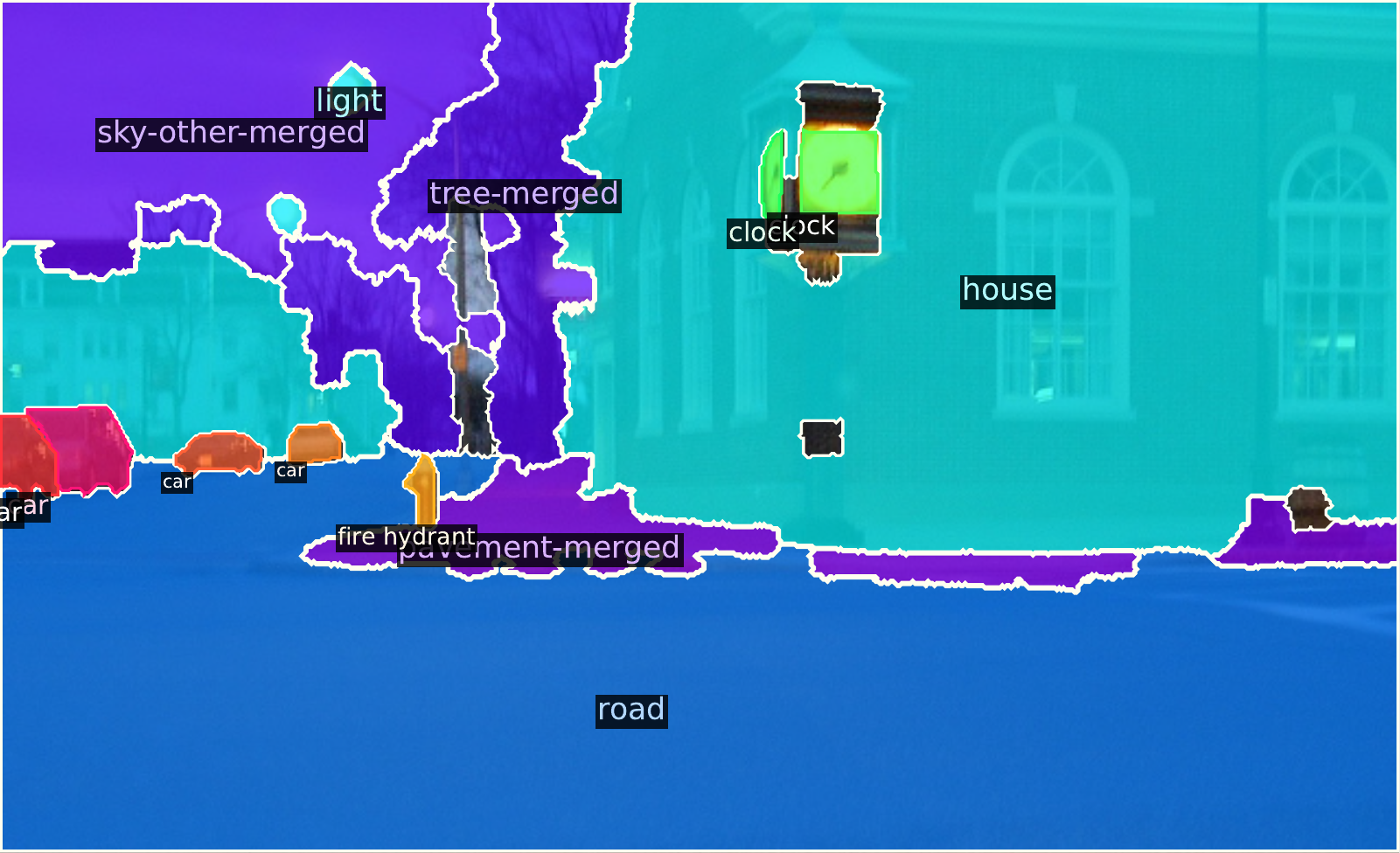}
\end{subfigure}
\begin{subfigure}[b]{\qual_width}
    \includegraphics[width=\textwidth]{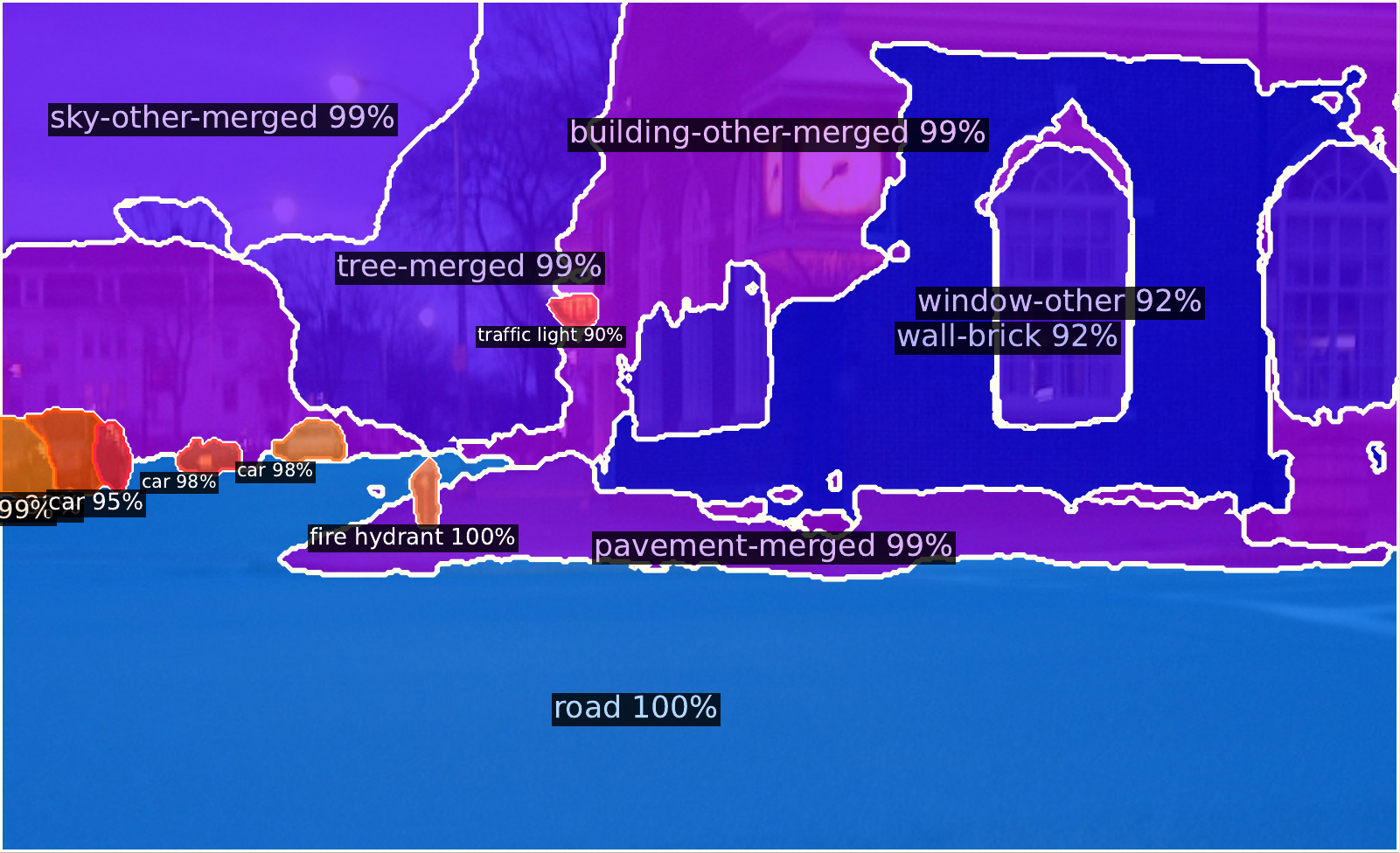}
\end{subfigure}
\begin{subfigure}[b]{\qual_width}
    \includegraphics[width=\textwidth]{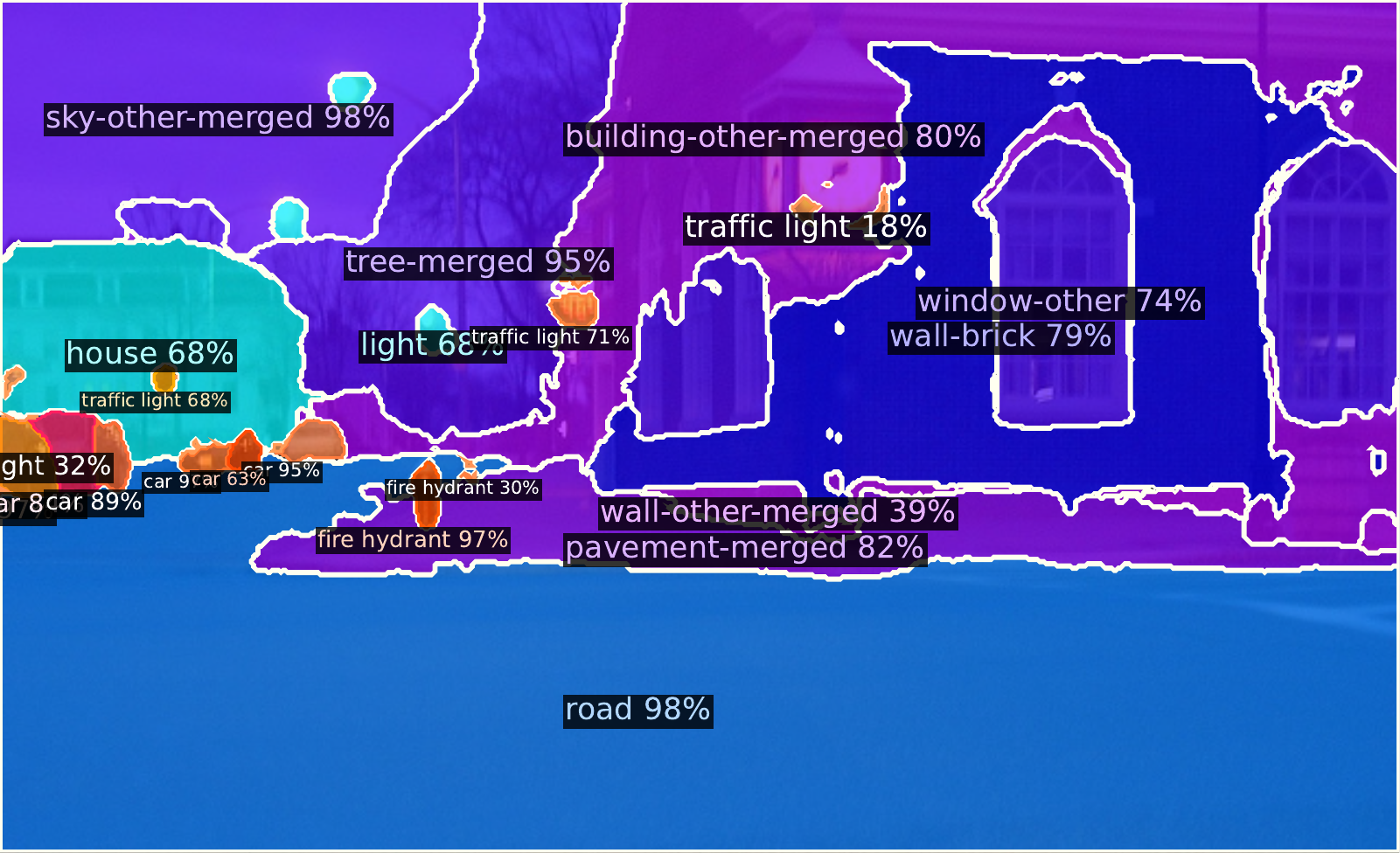}
\end{subfigure}
\begin{subfigure}[b]{\qual_width}
    \includegraphics[width=\textwidth]{Figures/NoUncertainty/COCO/RGB/000000186624.jpg}
\end{subfigure}
\begin{subfigure}[b]{\qual_width}
    \includegraphics[width=\textwidth]{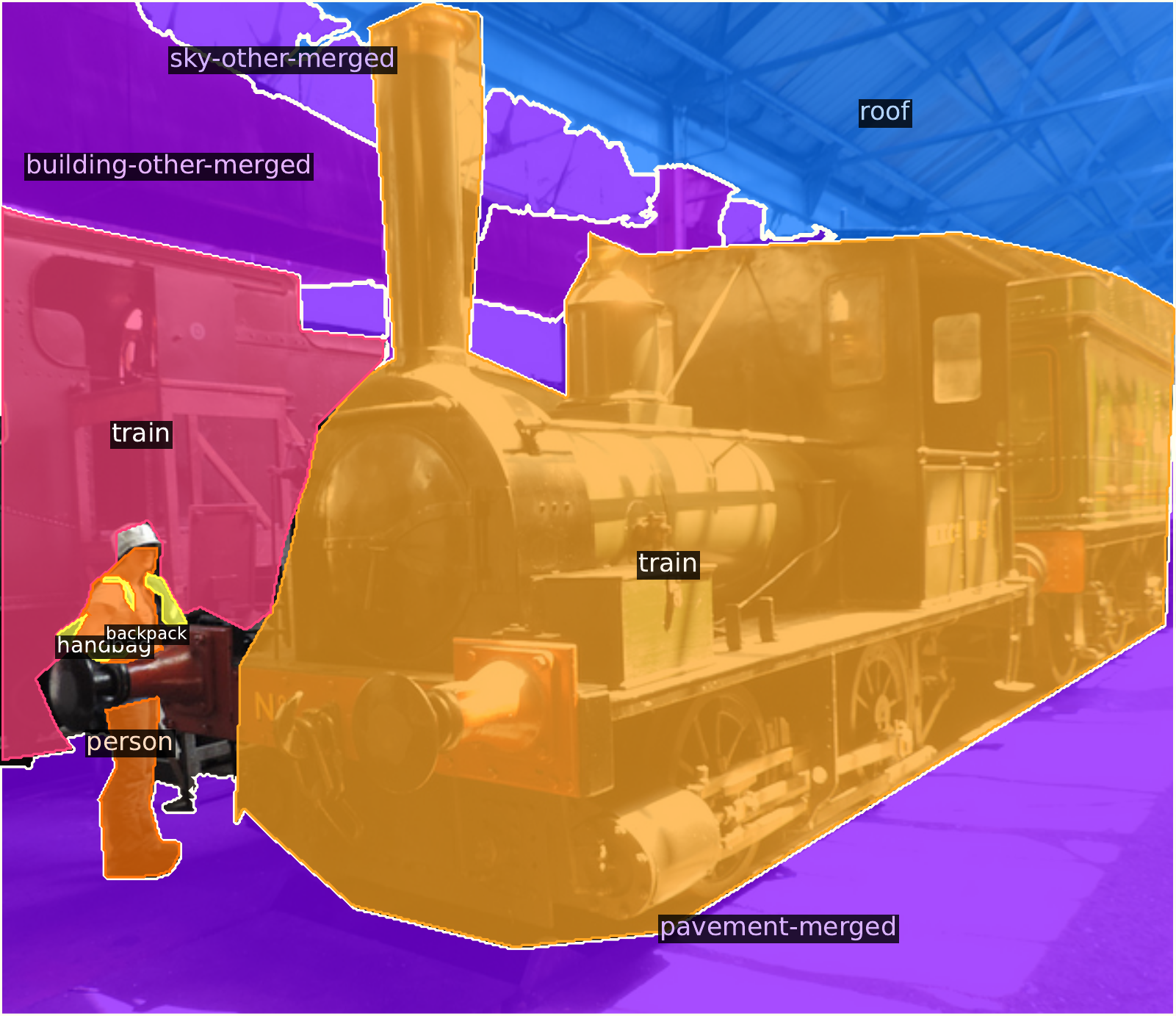}
\end{subfigure}
\begin{subfigure}[b]{\qual_width}
    \includegraphics[width=\textwidth]{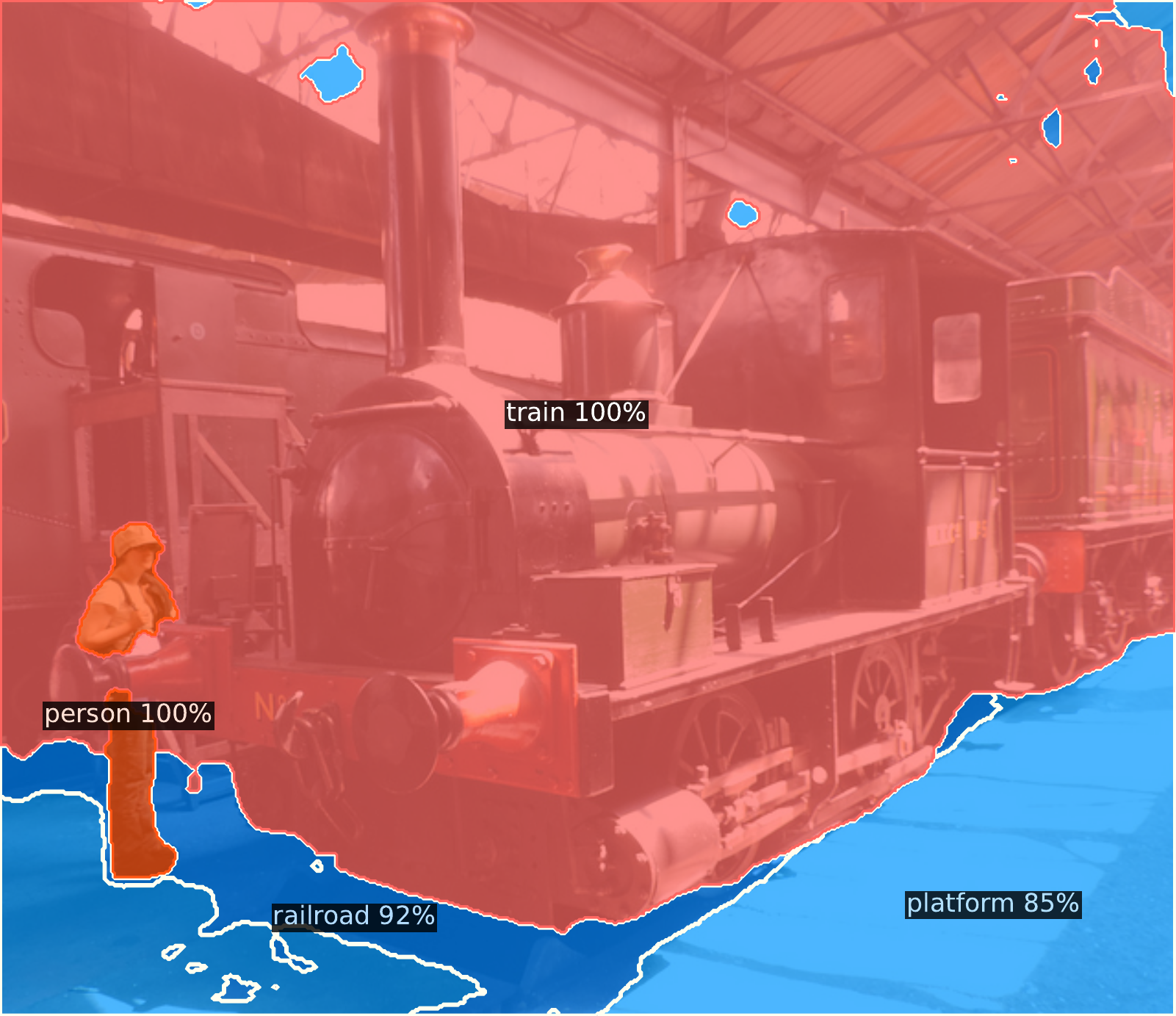}
\end{subfigure}
\begin{subfigure}[b]{\qual_width}
    \includegraphics[width=\textwidth]{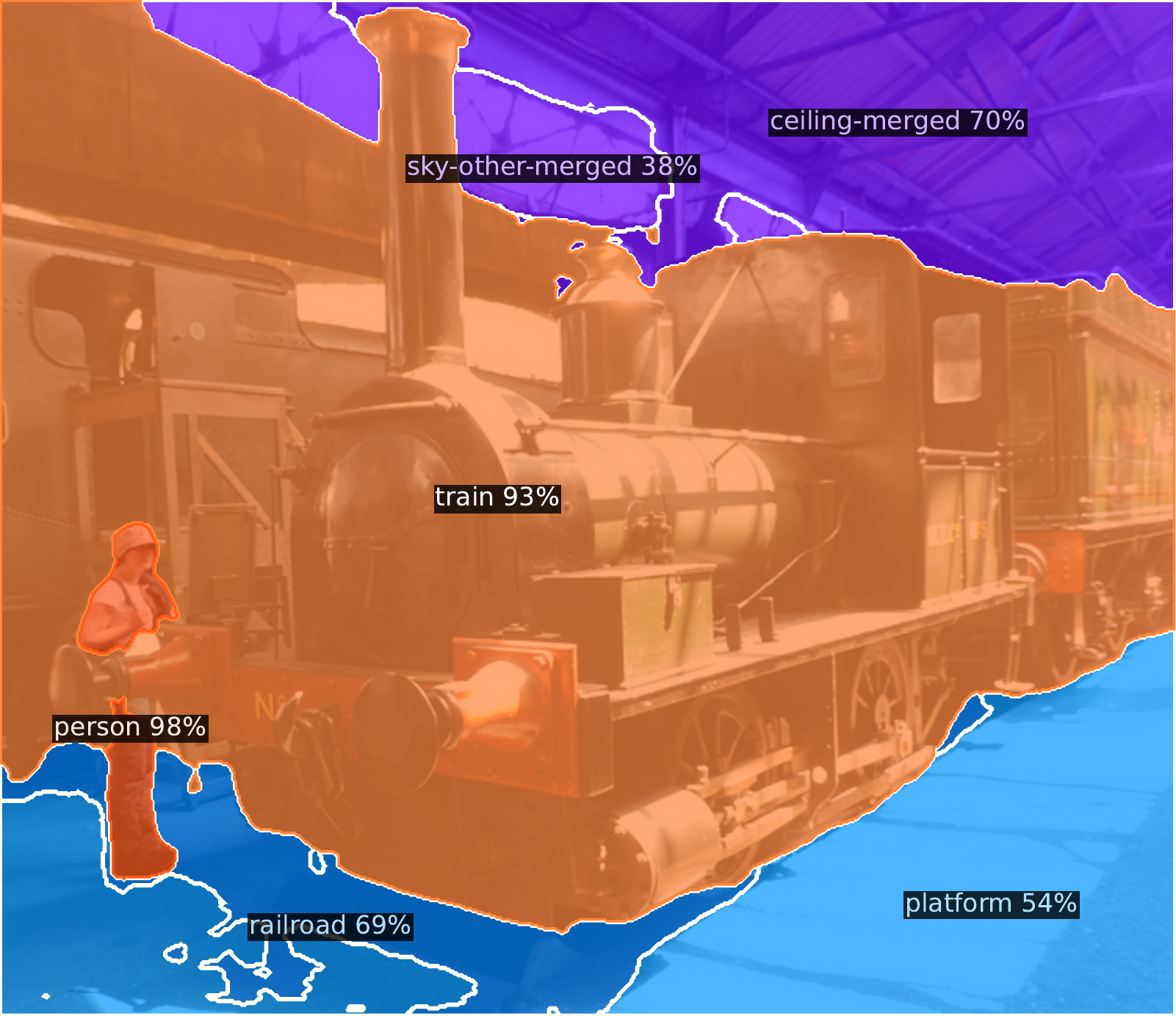}
\end{subfigure}
\begin{subfigure}[b]{\qual_width}
    \includegraphics[width=\textwidth]{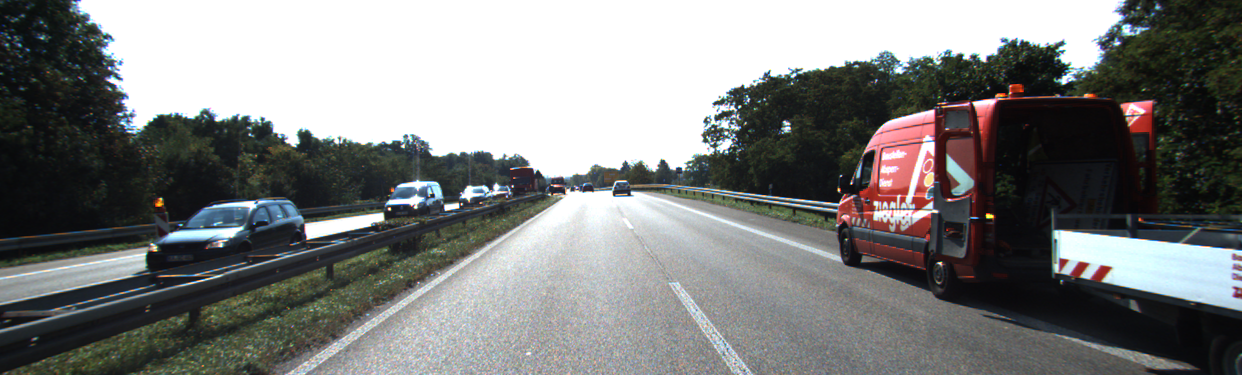}
\end{subfigure}
\begin{subfigure}[b]{\qual_width}
    \includegraphics[width=\textwidth]{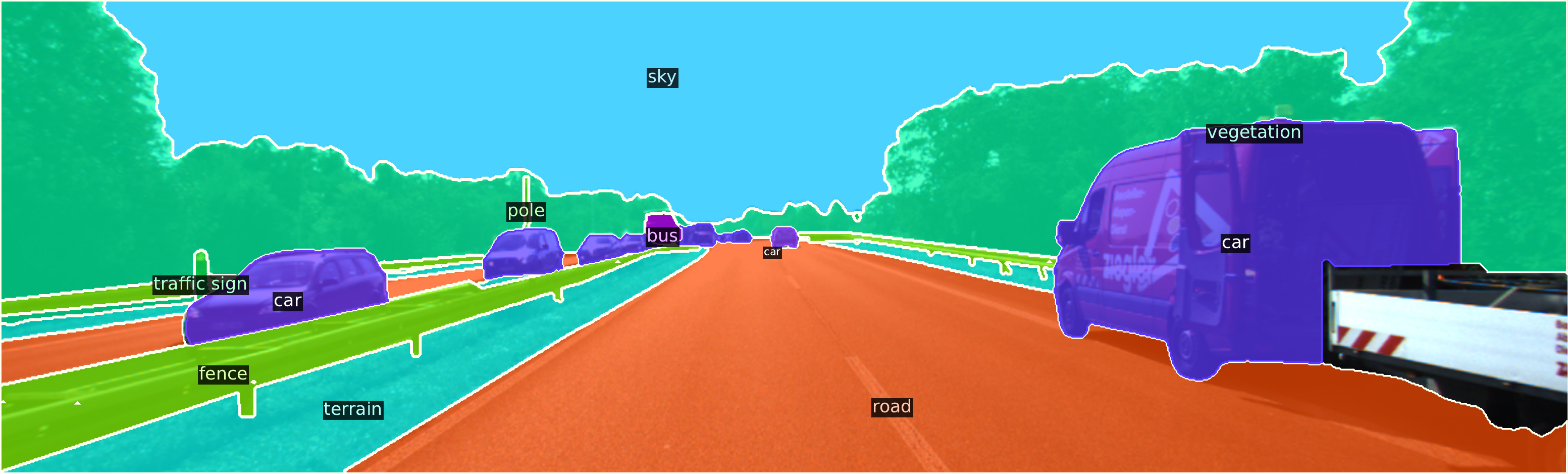}
\end{subfigure}
\begin{subfigure}[b]{\qual_width}
    \includegraphics[width=\textwidth]{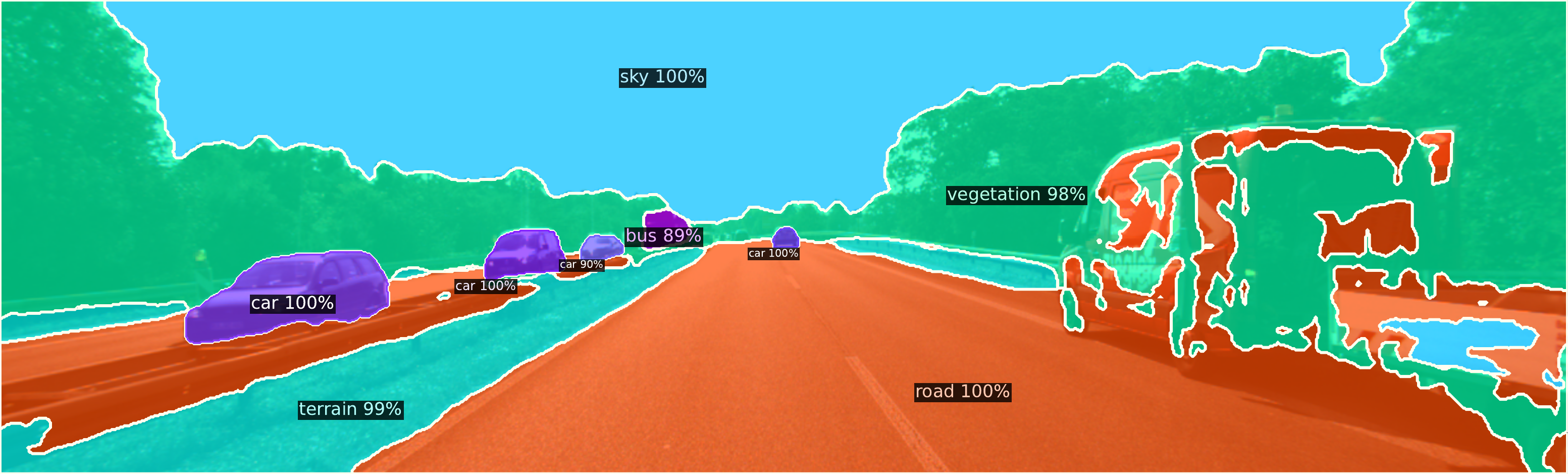}
\end{subfigure}
\begin{subfigure}[b]{\qual_width}
    \includegraphics[width=\textwidth]{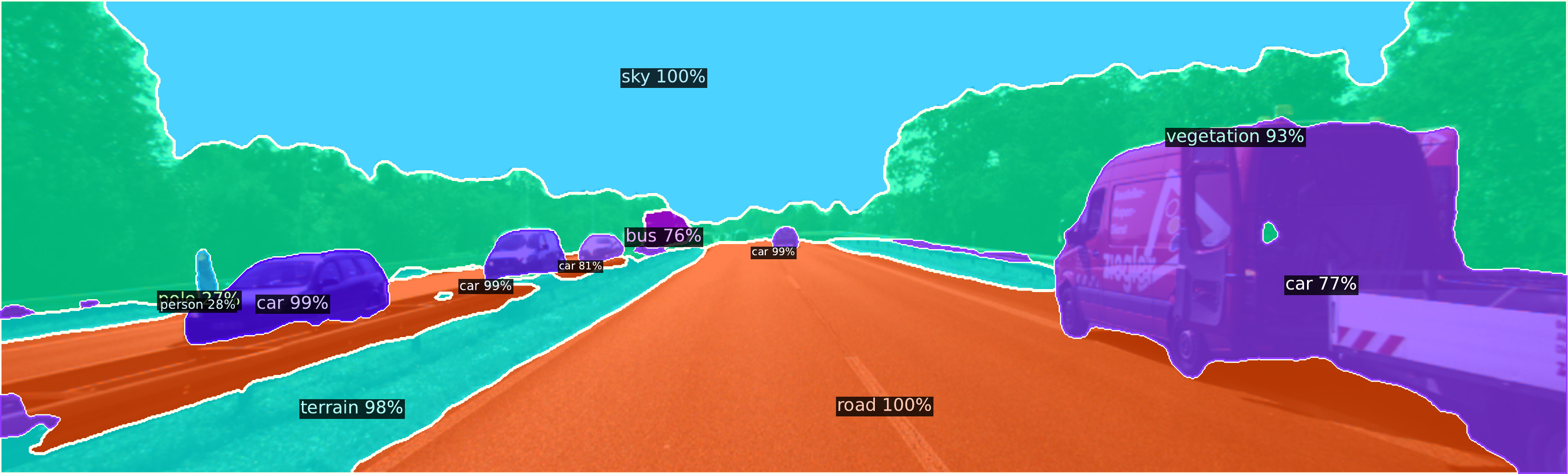}
\end{subfigure}
\begin{subfigure}[b]{\qual_width}
    \includegraphics[width=\textwidth]{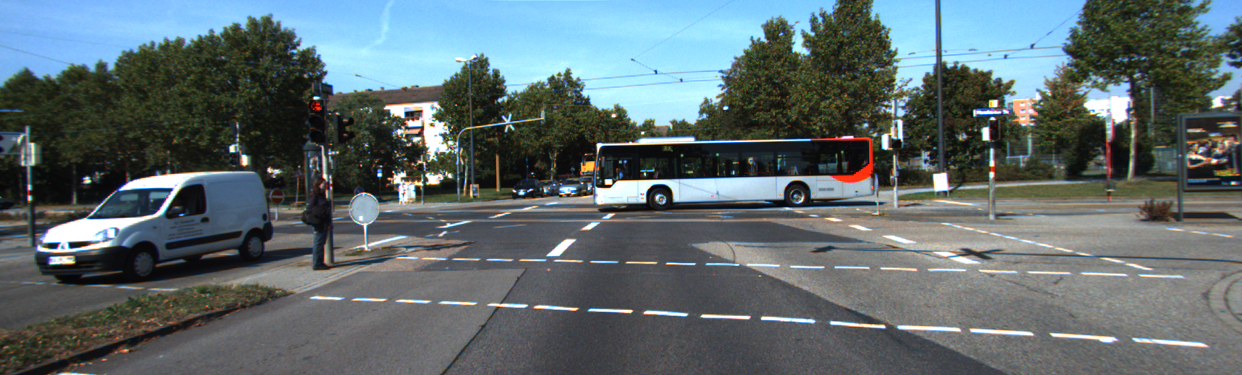}
\end{subfigure}
\begin{subfigure}[b]{\qual_width}
    \includegraphics[width=\textwidth]{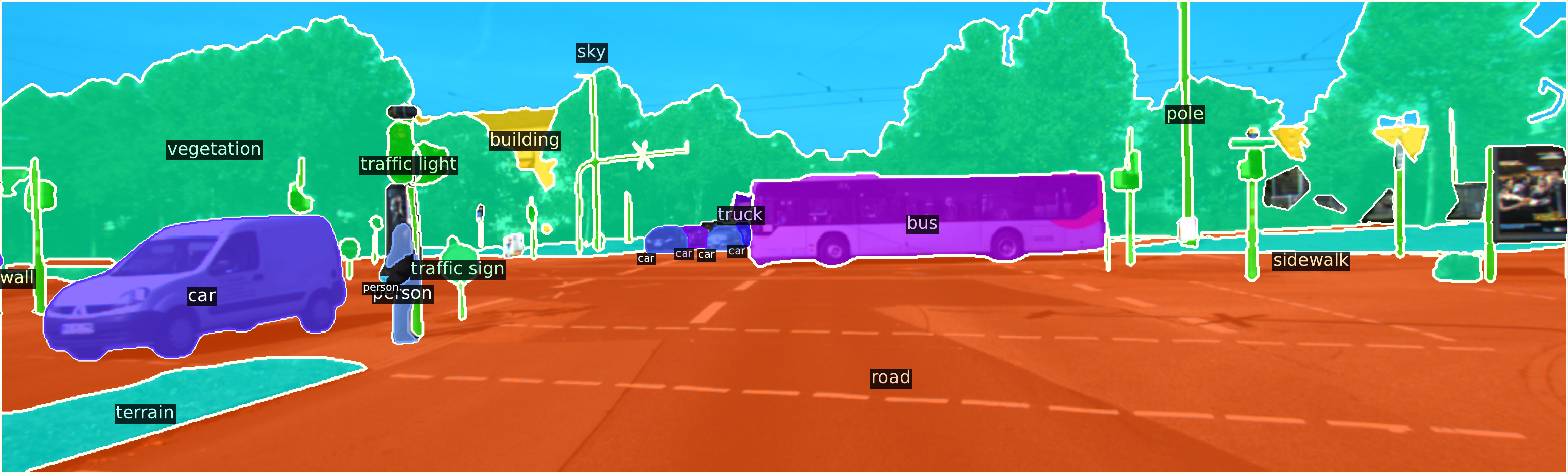}
\end{subfigure}
\begin{subfigure}[b]{\qual_width}
    \includegraphics[width=\textwidth]{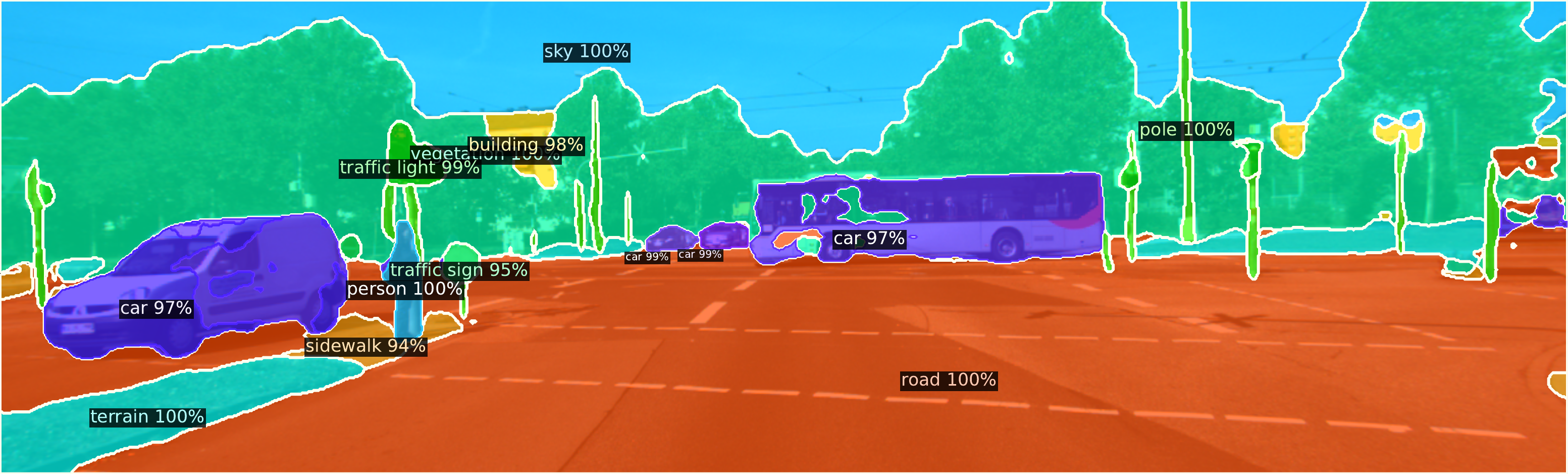}
\end{subfigure}
\begin{subfigure}[b]{\qual_width}
    \includegraphics[width=\textwidth]{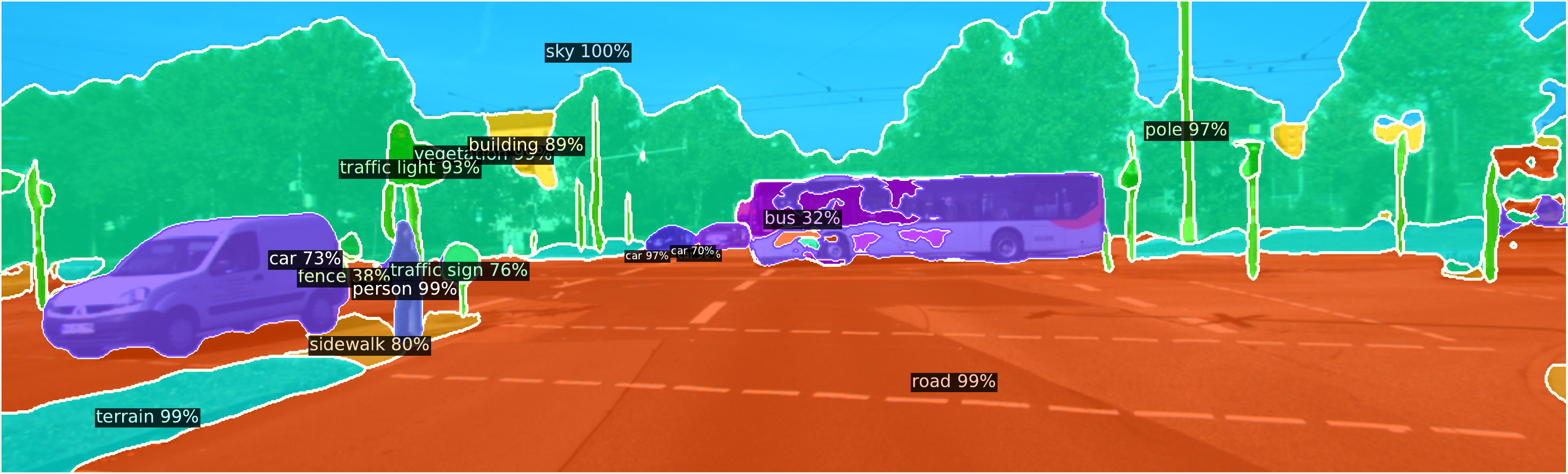}
\end{subfigure}
\begin{subfigure}[b]{\qual_width}
    \includegraphics[width=\textwidth]{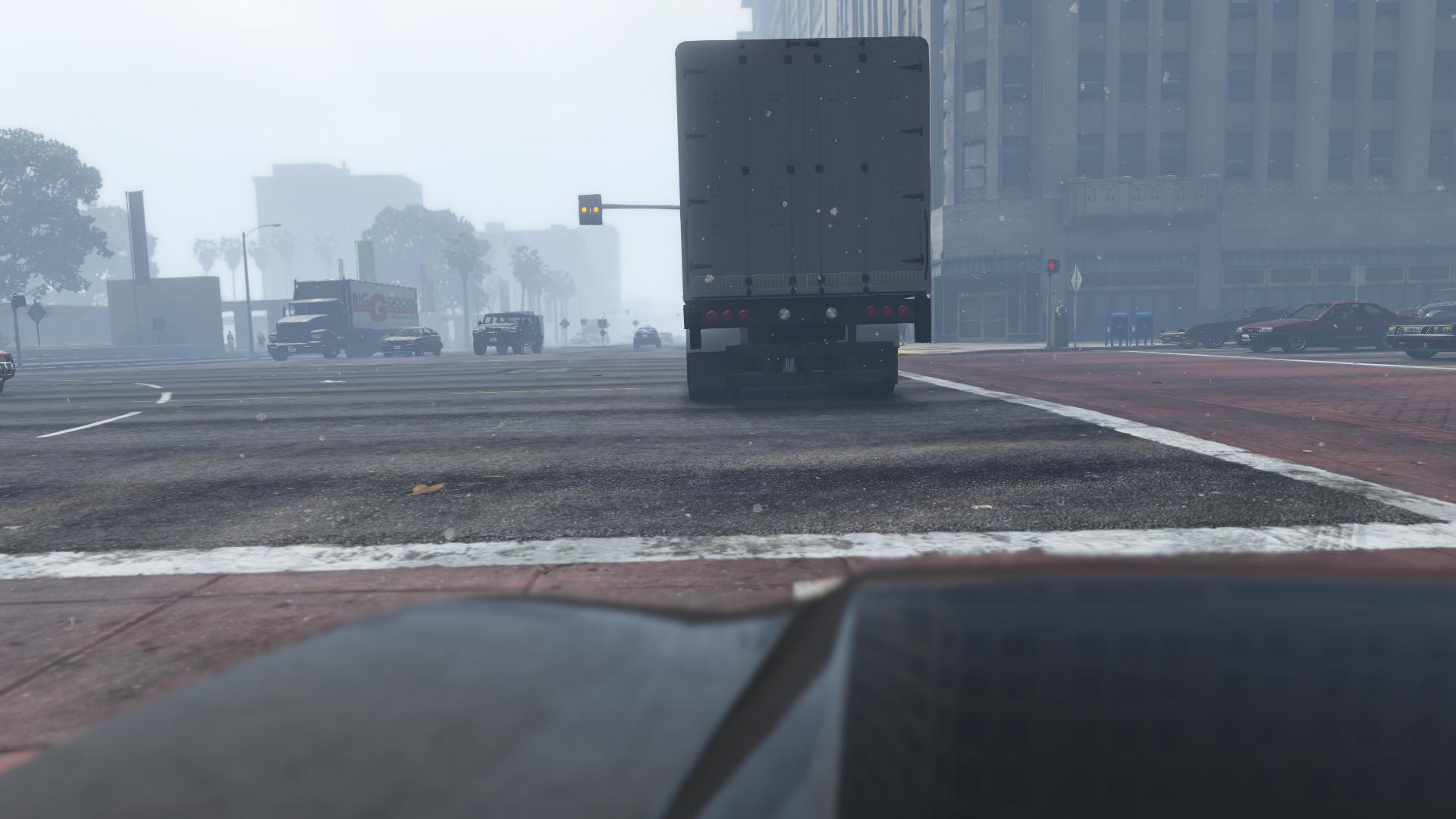}
\end{subfigure}
\begin{subfigure}[b]{\qual_width}
    \includegraphics[width=\textwidth]{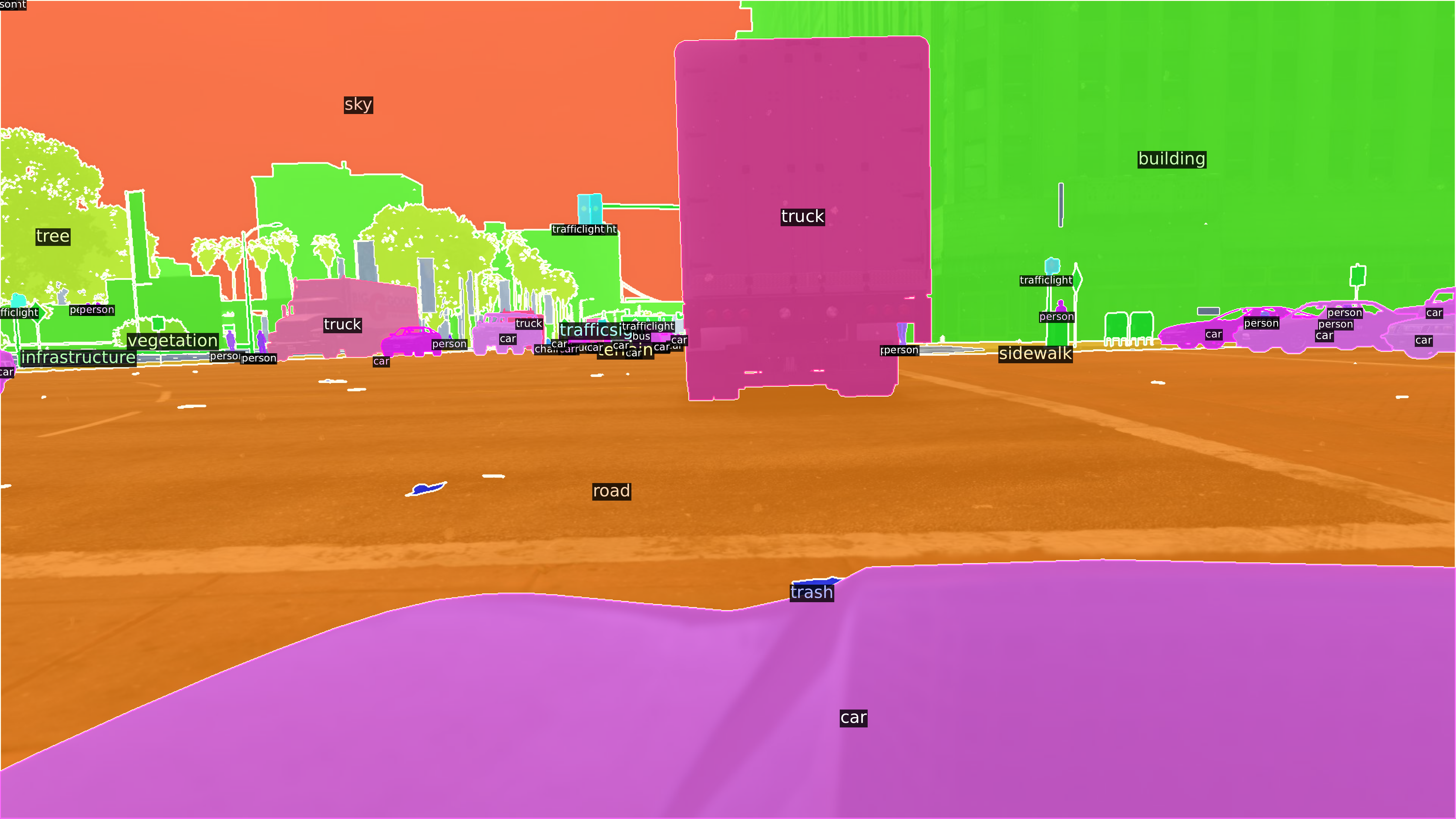}
\end{subfigure}
\begin{subfigure}[b]{\qual_width}
    \includegraphics[width=\textwidth]{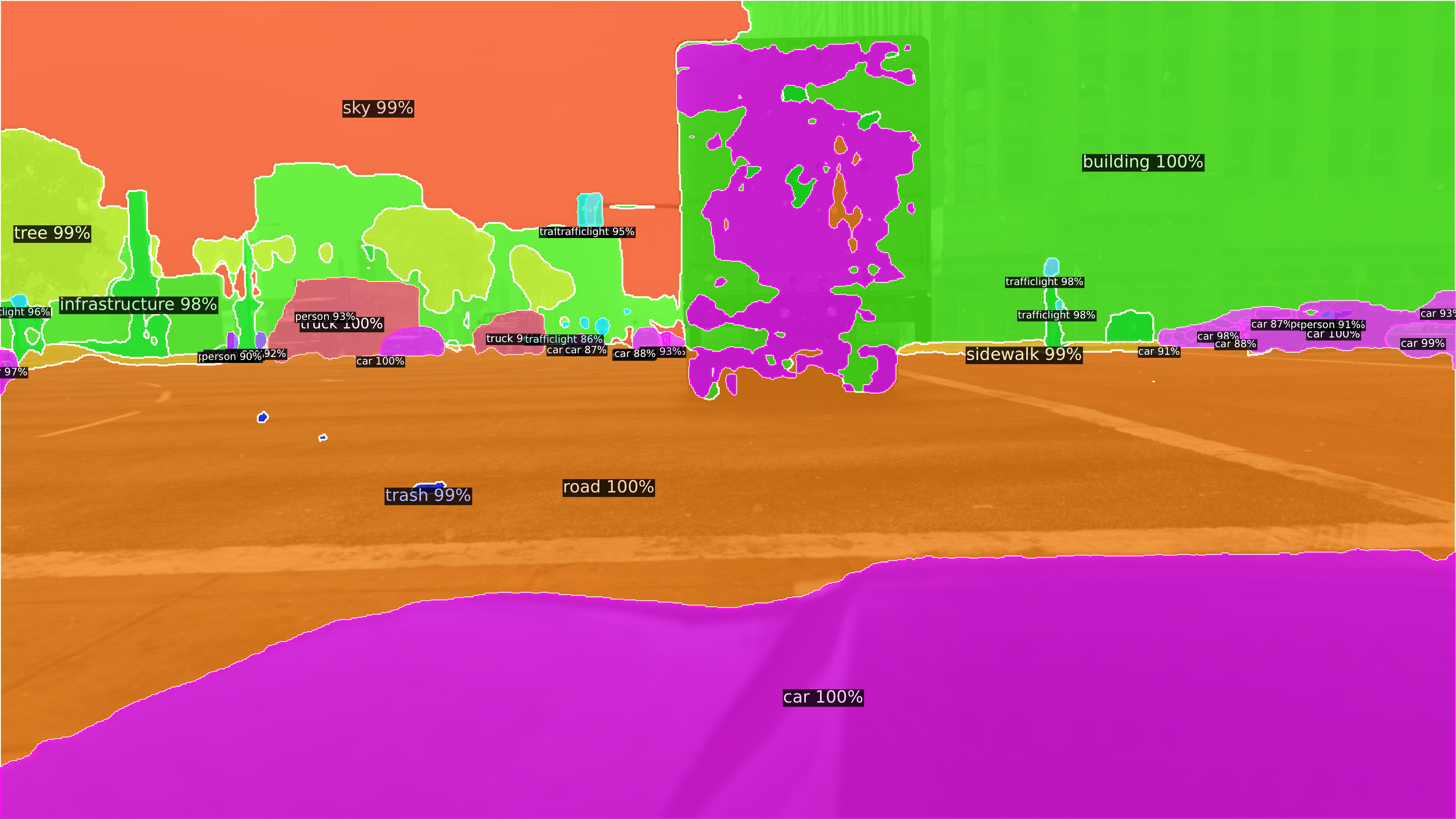}
\end{subfigure}
\begin{subfigure}[b]{\qual_width}
    \includegraphics[width=\textwidth]{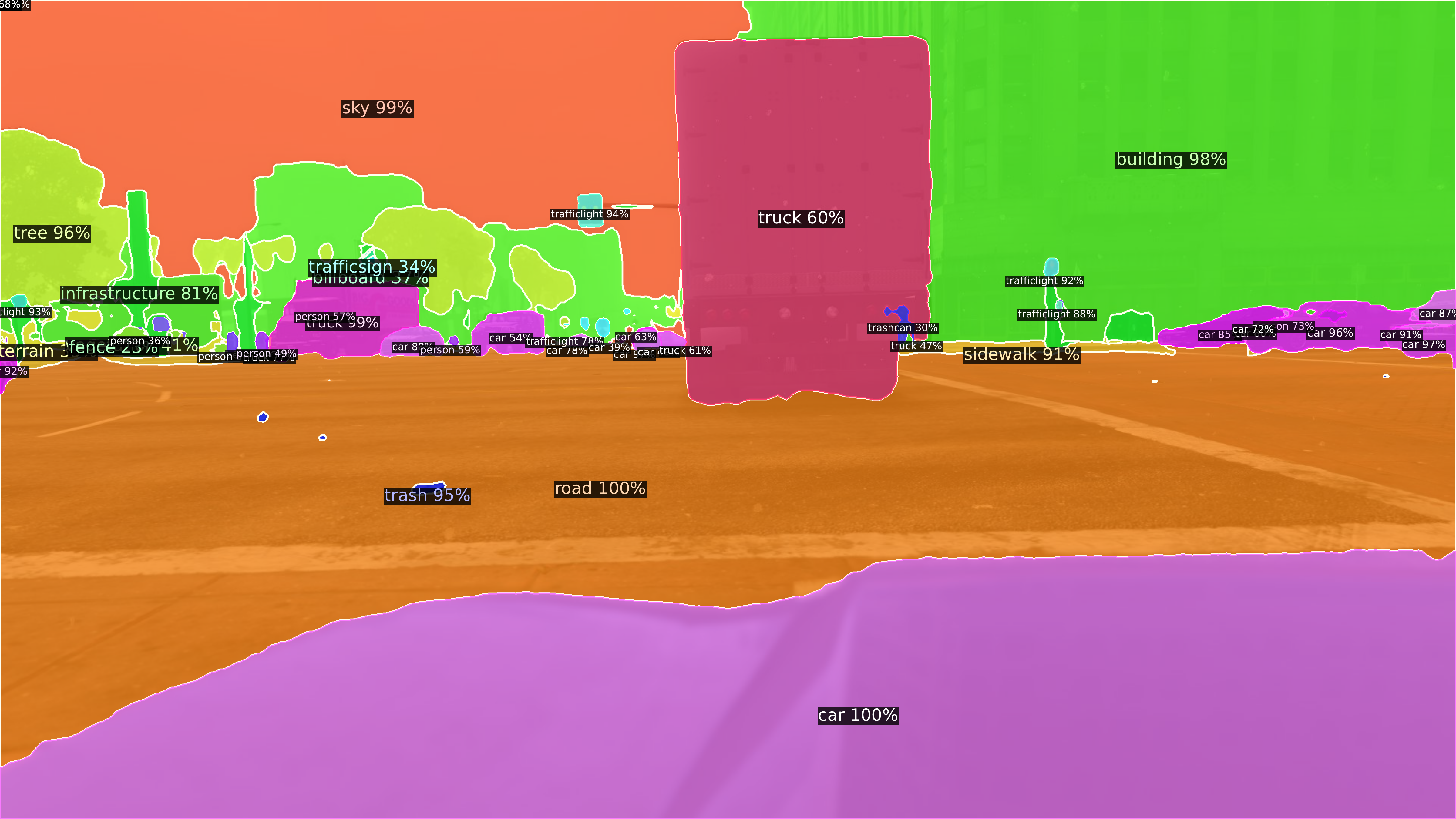}
\end{subfigure}
\begin{subfigure}[b]{\qual_width}
    \includegraphics[width=\textwidth]{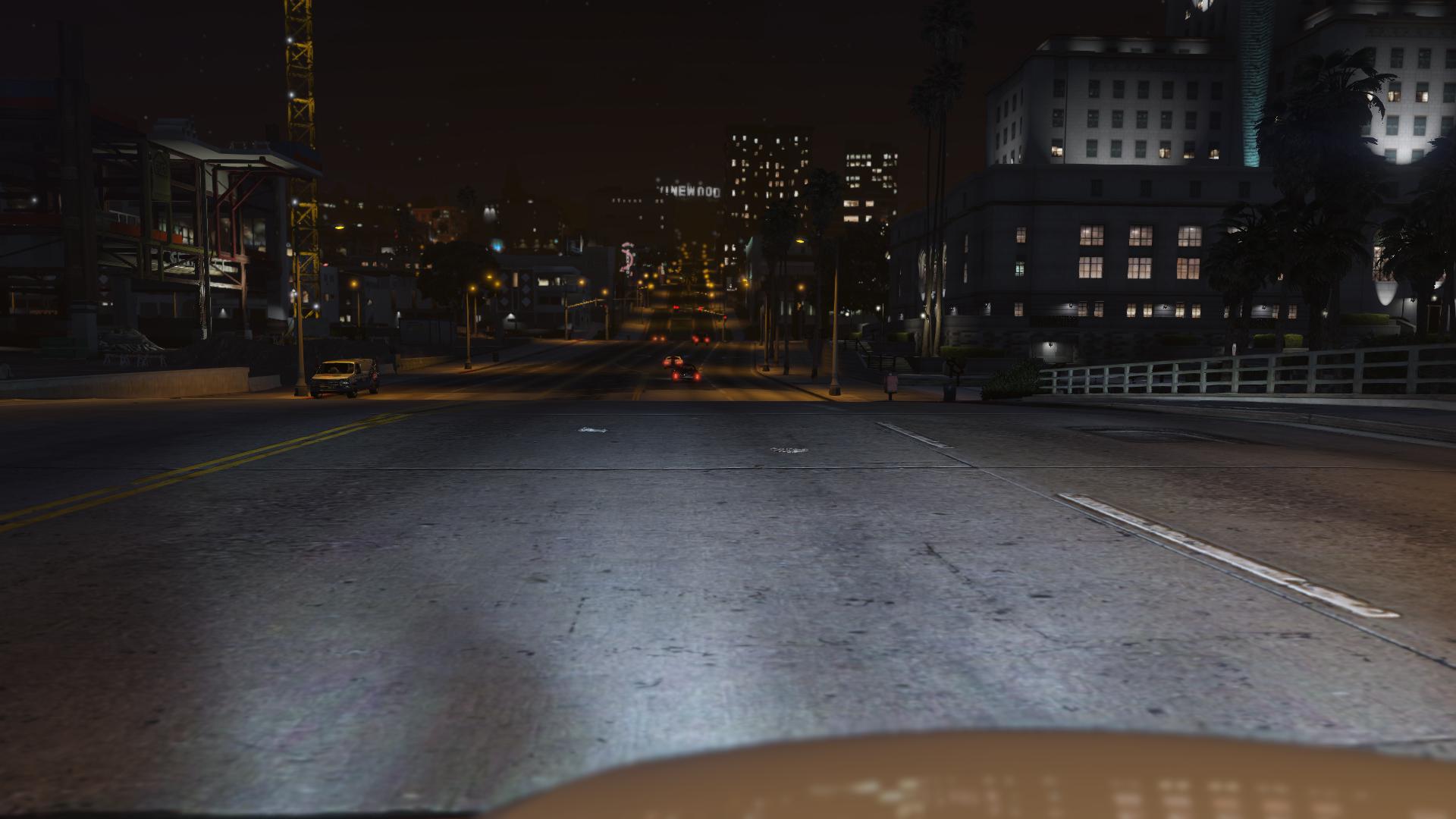}
\end{subfigure}
\begin{subfigure}[b]{\qual_width}
    \includegraphics[width=\textwidth]{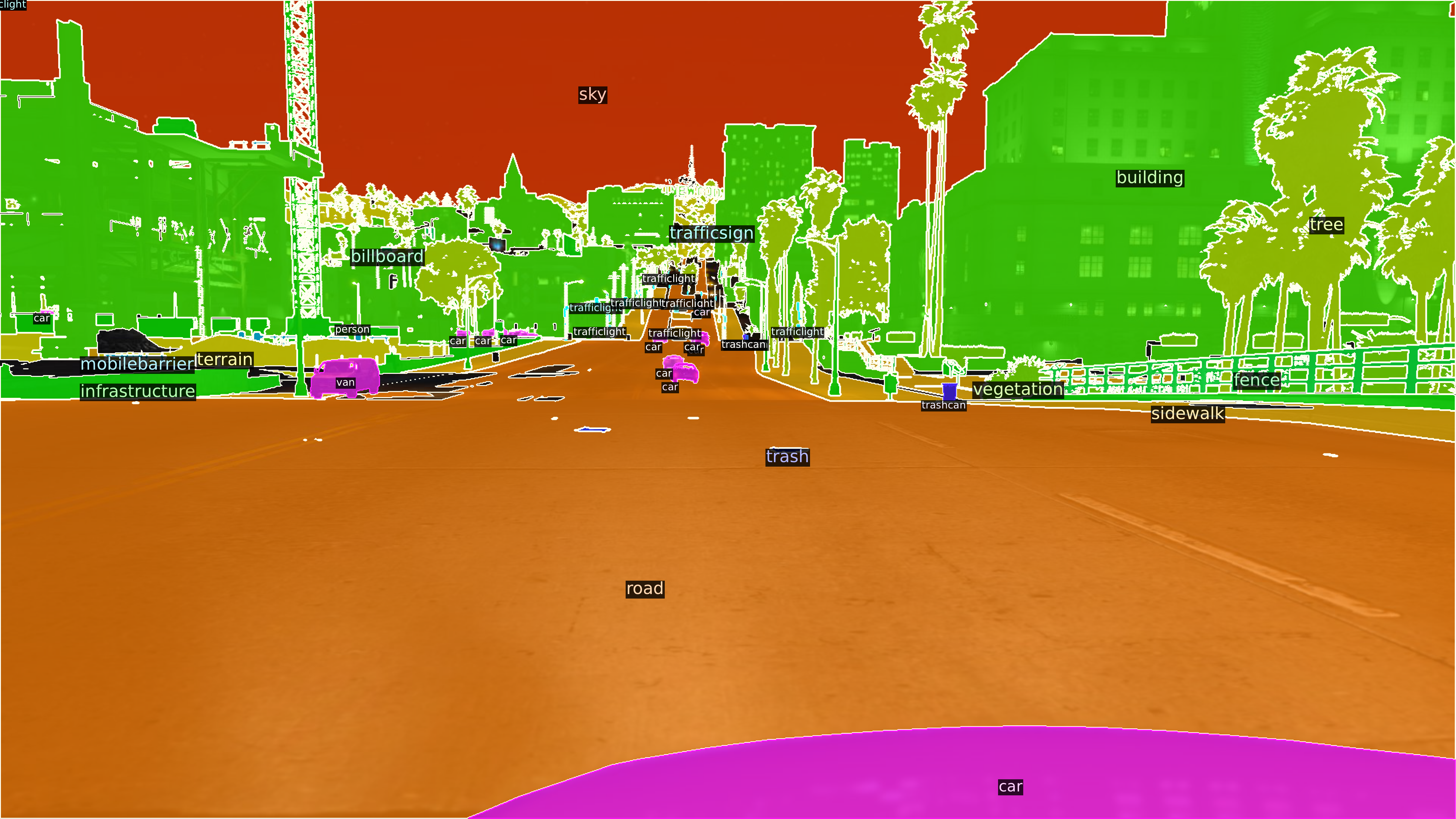}
\end{subfigure}
\begin{subfigure}[b]{\qual_width}
    \includegraphics[width=\textwidth]{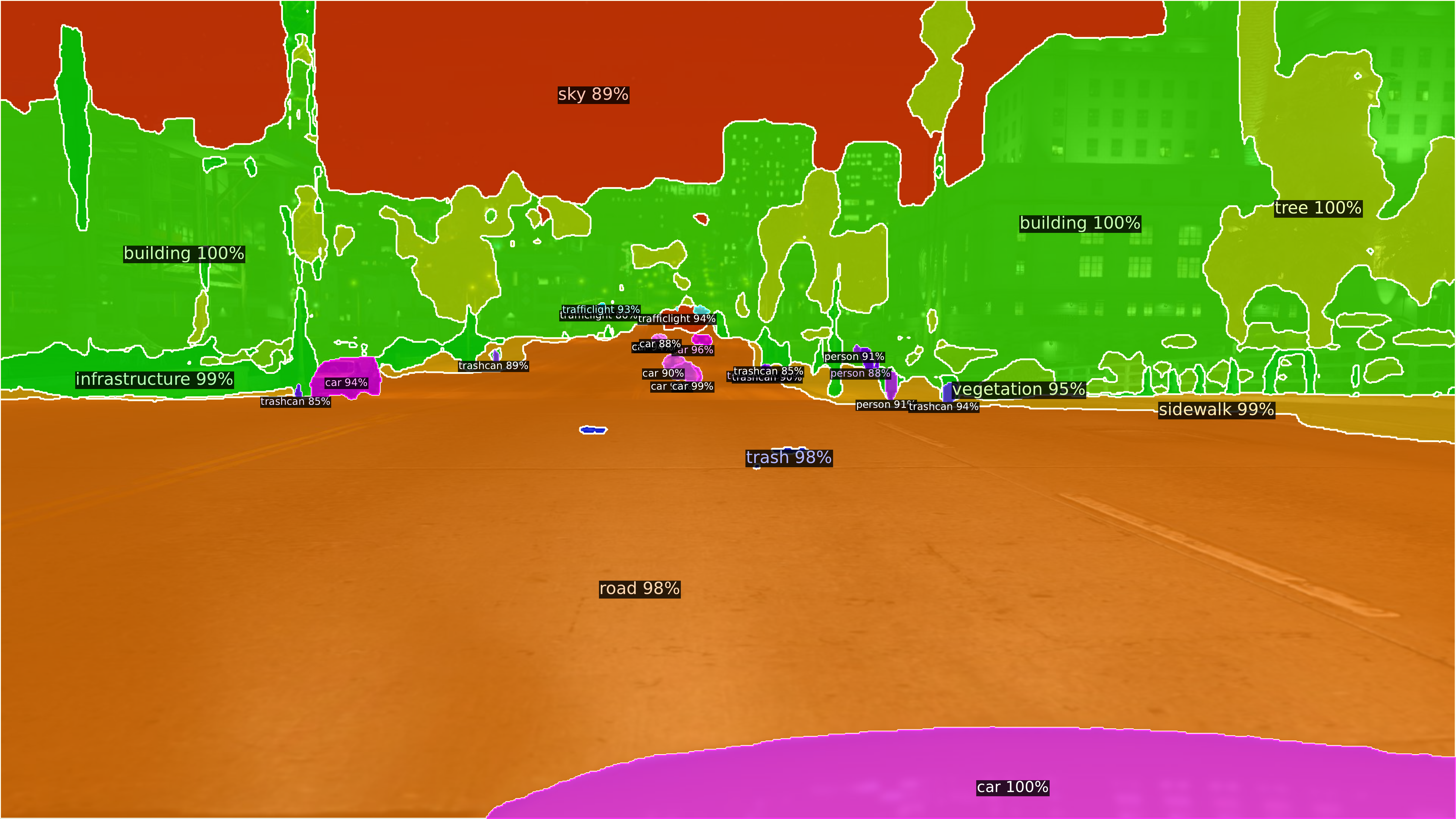}
\end{subfigure}
\begin{subfigure}[b]{\qual_width}
    \includegraphics[width=\textwidth]{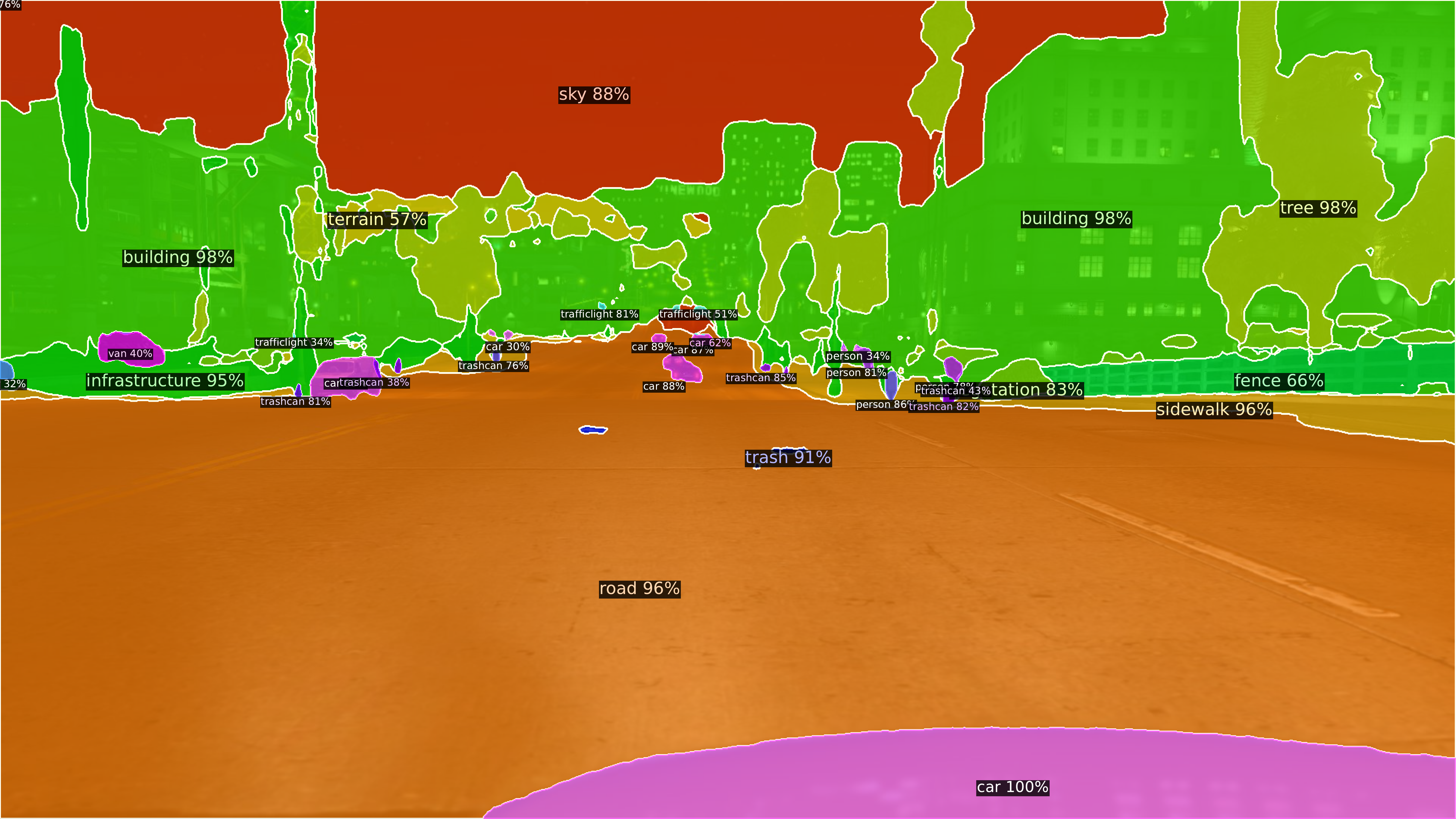}
\end{subfigure}
  \caption{Qualitative examples from the COCO (top 2 rows), KITTI-STEP (middle 2 rows) and VIPER (bottom 2 rows) validation sets. Left to right: input image, ground truth, baseline segmentation, our approach with 15 Monte Carlo samples.}
  \label{fig:qualitative_examples}
\end{figure}

In \cref{fig:qualitative_examples} we examine selected images and their panoptic predictions for both datasets. Broadly speaking, we see that our approach is able to slightly improve the segmentation versus the baseline as reflected in our quantitative experiments. In some cases in the COCO dataset, however, ground truth ambiguity may result in some improvements not being reflected in the PQ metric.

\begin{figure}
\def\qual_ent_width{0.48\textwidth}
\centering
\begin{subfigure}{\qual_ent_width}
    \includegraphics[width=\textwidth]{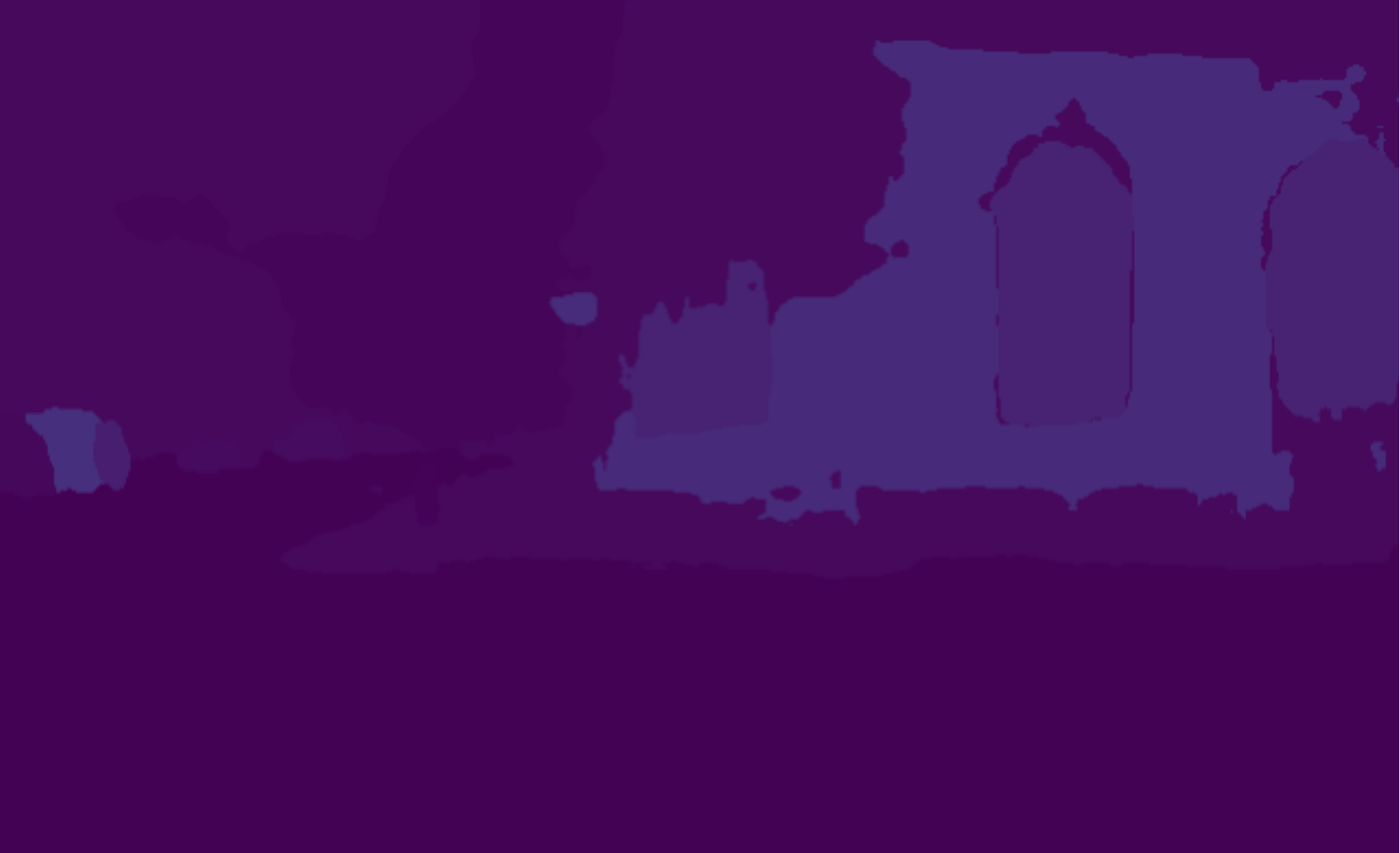}
\end{subfigure}
\begin{subfigure}{\qual_ent_width}
    \includegraphics[width=\textwidth]{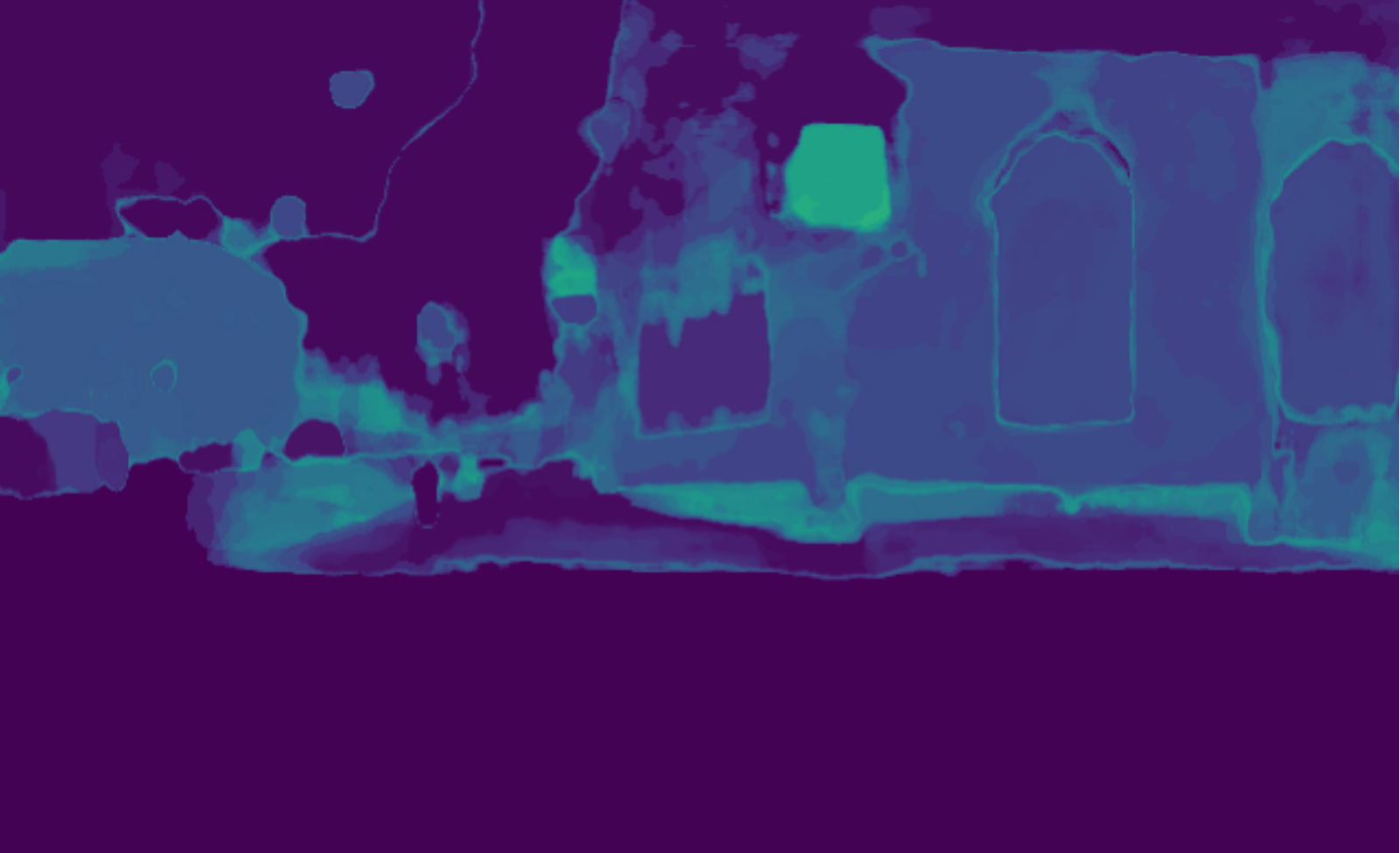}
\end{subfigure}
\caption{Side-by-side comparisons of predictive entropy uncertainty heatmaps for the baseline (left) and our approach with 15 MC samples (right) for the top-most example shown in \cref{fig:qualitative_examples}. Lighter shades indicate greater uncertainty.}
\label{fig:uncertainty_heatmap}
\end{figure}

In \cref{fig:uncertainty_heatmap}, we provide heatmaps comparing the softmax and predictive entropy for baseline models and our approach. This qualitatively demonstrates the superior uncertainty estimates made possible through our approach, as we see increased heterogeneity in many areas, indicating a more diverse set of predictions have been considered. Given our quantitative results in \cref{fig:pred_entropy_hist}, these qualitative differences are to be expected.

\section{Conclusion}
We present an algorithm that allows ensemble uncertainty estimates to, for the first time, be used in the panoptic segmentation domain. We demonstrate its effectiveness with Monte Carlo Dropout \cite{gal_dropout_2016} and the transformer-based DETR network \cite{carion_end--end_2020} specifically, with which we generate panoptic segmentation results that improve slightly over the baseline DETR and more significantly over the Hungarian algorithm. Most importantly, we can generate more reliable confidence measures using mathematically grounded uncertainty functions such as the predictive entropy, subject to the limits of the chosen sample generation process such as MC Dropout in our experiments. This allows errors under various scenarios to be identified and handled appropriately. Our approach is reasonably straightforward and fairly broad in its applicability. As presented, existing neural network models can be used without any architecture modification or retraining. More generally, our approach should be compatible with any ensemble-based uncertainty estimation process (\textit{e.g.} \cite{lakshminarayanan_simple_2017}) and most recent universal panoptic segmentation architectures including the current state of the art \cite{li_panoptic_2022}. We hope that these findings encourage future work on obtaining better uncertainty estimates for panoptic segmentation, as they can now be fairly compared to known approaches such as Monte Carlo Dropout \cite{gal_dropout_2016} among others.

\section{Acknowledgements}
We would like to thank Nvidia for providing one of the GPUs used for our experiments.

\bibliographystyle{splncs04}
\bibliography{main}

\clearpage

\section*{Supplementary Material}
\appendix
\renewcommand{\thefigure}{A\arabic{figure}}
\renewcommand{\thealgocf}{A\arabic{algocf}}
\renewcommand{\thetable}{A\arabic{table}}
\setcounter{figure}{0}
\setcounter{algocf}{0}
\setcounter{table}{0}

\section{Panoptic Segmentation Visualization}
\label{app:sec:panoptic_viz}
\begin{figure}
    \centering
    \begin{subfigure}[b]{0.48\textwidth}
         \centering
         \includegraphics[width=\textwidth]{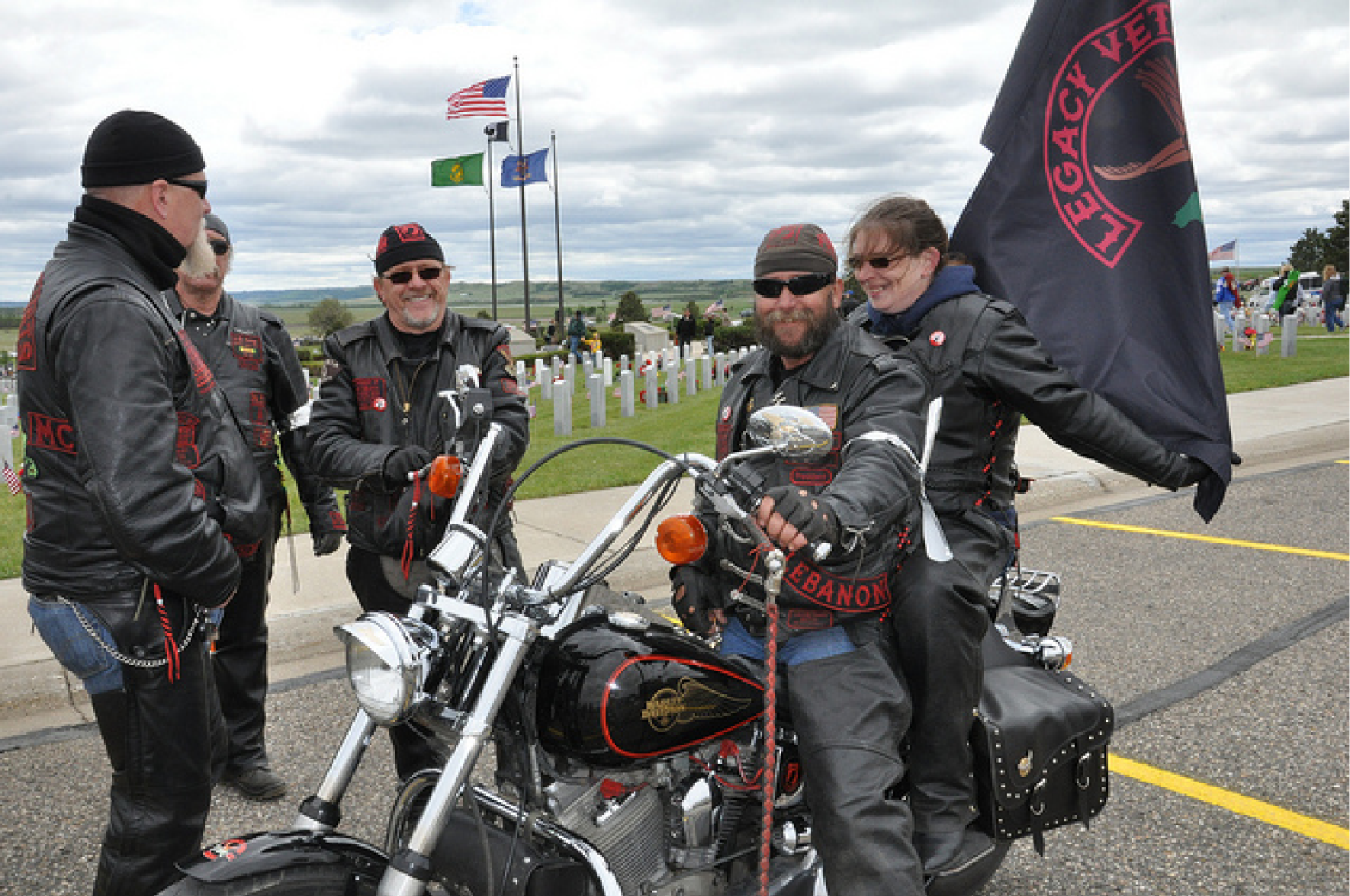}
         \caption{RGB Image}
         \label{app:fig:rgbimg}
    \end{subfigure}
    \begin{subfigure}[b]{0.48\textwidth}
        \centering
         \includegraphics[width=\textwidth]{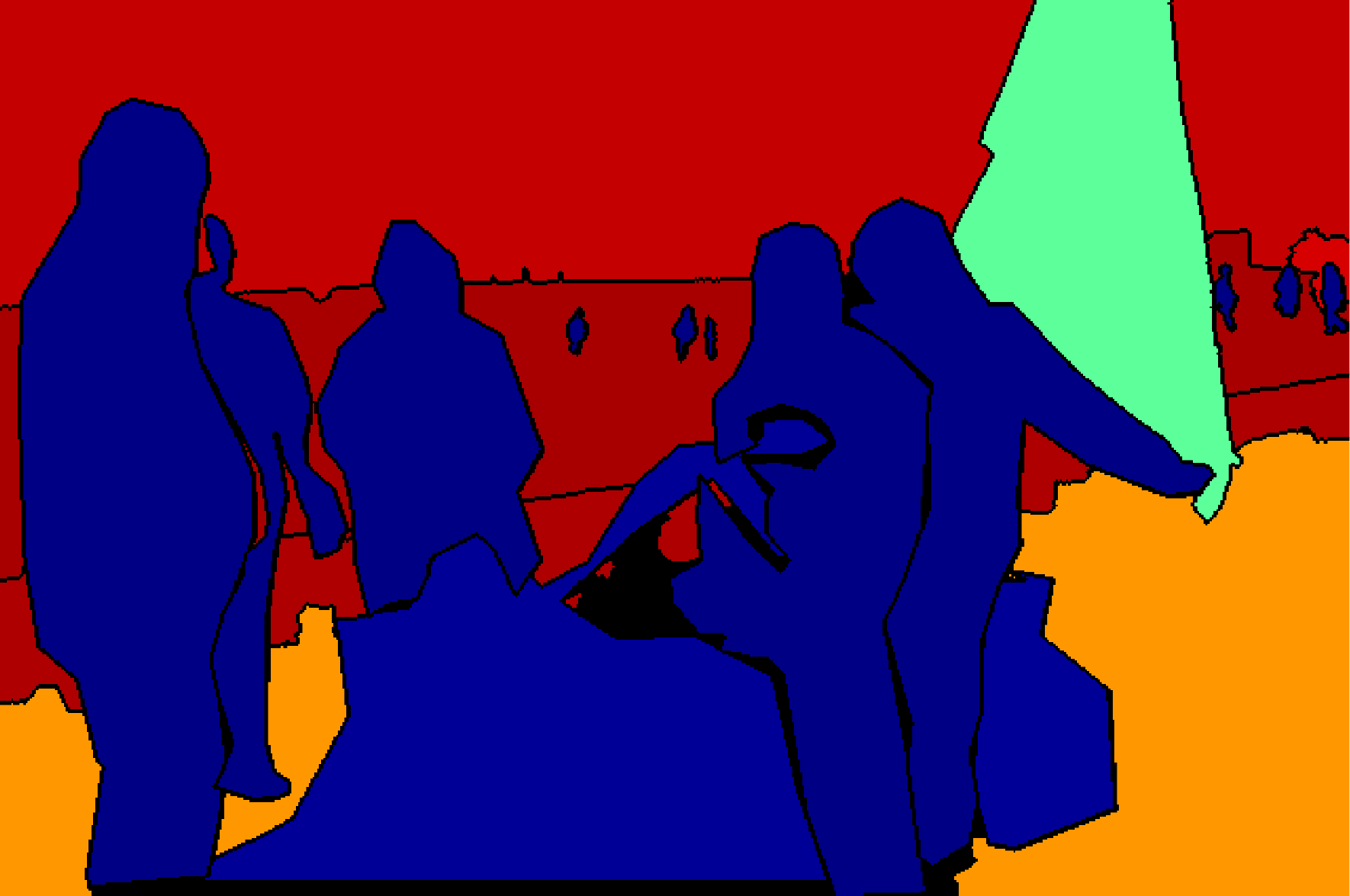}
         \caption{Semantic Segmentation}
        \label{app:fig:semanticexample}
    \end{subfigure}
    \begin{subfigure}[b]{0.48\textwidth}
         \centering
         \includegraphics[width=\textwidth]{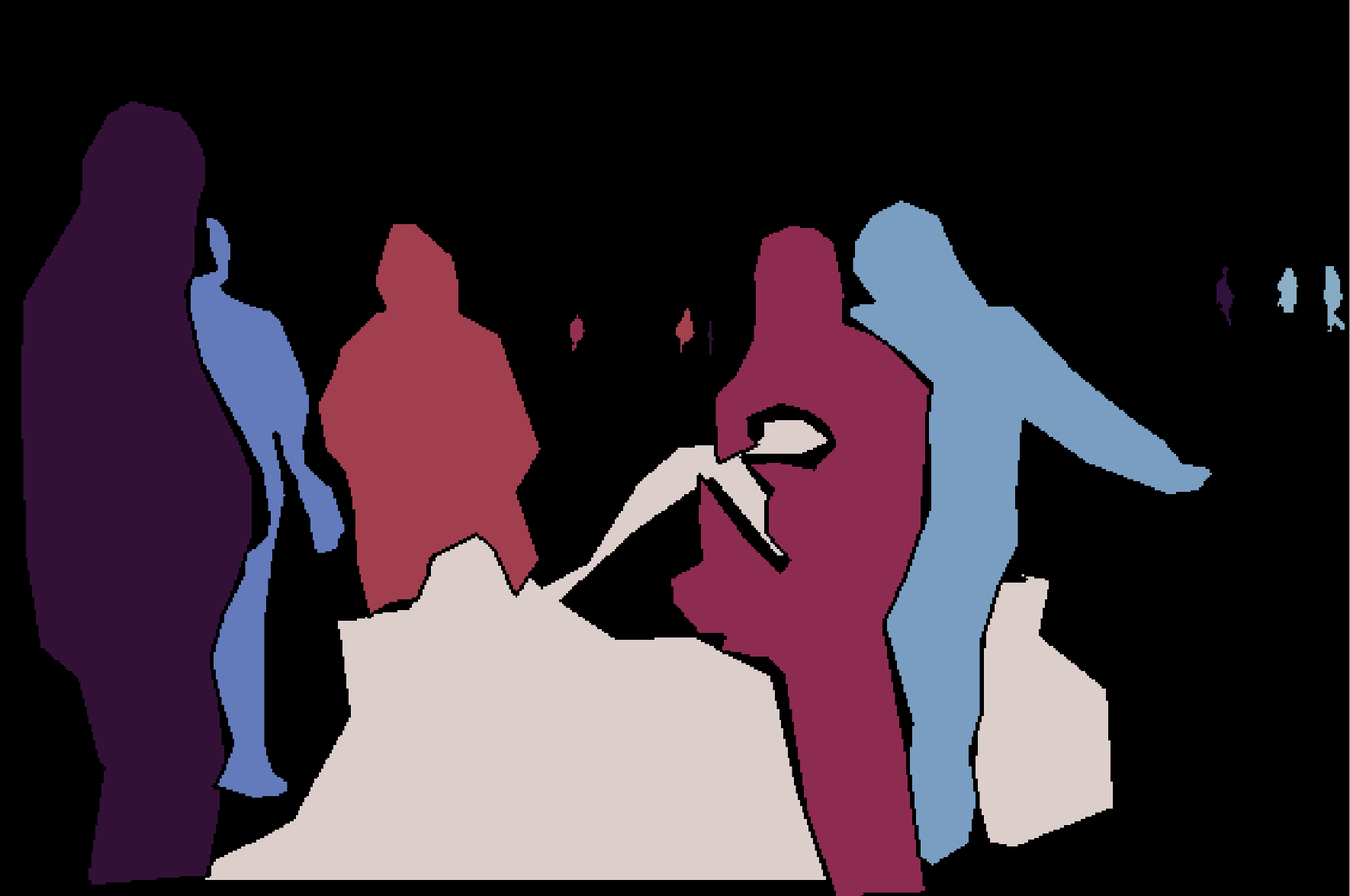}
         \caption{Instance Segmentation}
         \label{app:fig:thingexample}
    \end{subfigure}
    \begin{subfigure}[b]{0.48\textwidth}
        \centering
         \includegraphics[width=\textwidth]{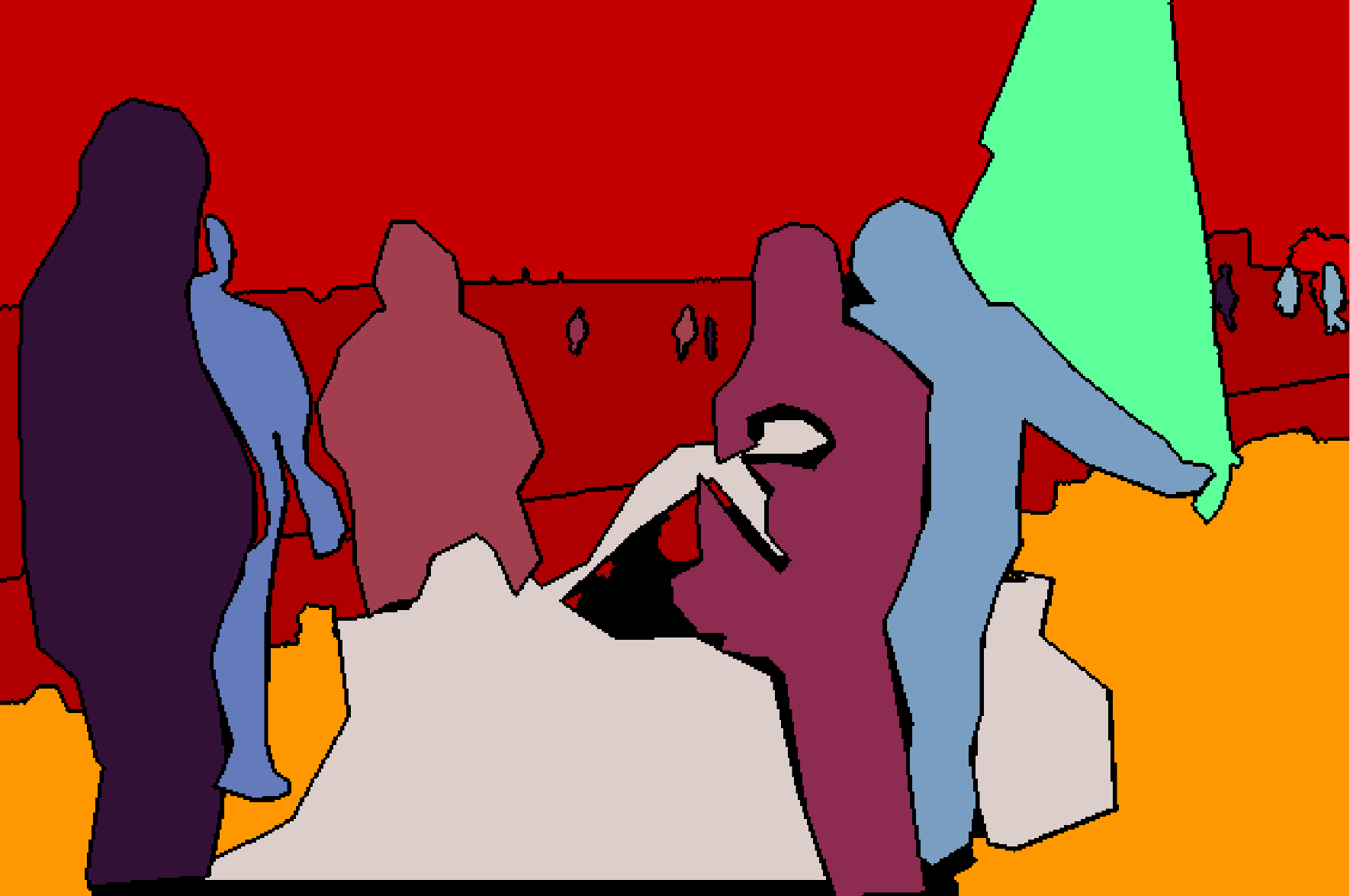}
         \caption{Panoptic Segmentation}
        \label{app:fig:bothexample}
    \end{subfigure}
    \captionsetup{subrefformat=parens}
    \caption{Comparison of semantic, instance and panoptic segmentation, with ground truth data for an image from the COCO dataset \cite{lin_microsoft_2014}. Semantic segmentation \subref{app:fig:semanticexample} uses per-pixel class labels. Instance segmentation \subref{app:fig:thingexample} uses per-object mask and class label. Panoptic segmentation \subref{app:fig:bothexample} uses per-pixel class and instance labels. Inspired by Fig. 1 in \cite{kirillov_panoptic_2019}.}
    \label{app:fig:panoptic_example}
\end{figure}

In \cref{app:fig:panoptic_example}, we illustrate the difference between panoptic, semantic and instance segmentation. Note that the image from the COCO dataset \cite{lin_microsoft_2014} is generally representative of the complexity of the scenes in the datasets we use, with in this particular case a fair amount of similarity between objects and people in the scene. This complexity in the dataset, combined with the complexity inherent in the task of panoptic segmentation, helps explain why to our knowledge all panoptic segmentation algorithms use neural networks in some fashion.

\section{Algorithm details}
With most algorithms featured in the paper, we formally describe the remaining algorithm here in \cref{app:alg:perSampleSegmentation}. This expands on the overview provided in the paper.

\begin{algorithm}
\DontPrintSemicolon
\caption{perSampleSeg(\textit{L}, \textit{M})}
\label{app:alg:perSampleSegmentation}
\SetKwFunction{upsc}{upscale}
\KwIn{Logits $L^{Q \times N \times C}$, Masks $M^{Q \times N \times h \times w}$}
\KwOut{Per-sample Proposal Map $S^{Q \times H \times W}$, List length $Q$, of proposal softmax vectors $I^{K \times C}$}
\tcp{This algorithm based on the baseline approach \cite{carion_end--end_2020}}
\ForEach{Sample $q \in Q$}{
    $S \leftarrow \emptyset$\;
    \ForEach{proposal $n \in N \subseteq q$}{
            $l^{C} \leftarrow \soft L_{q, n}$\;

            label $\leftarrow \argmax_{C} l$\;
            \If{label = background}{remove proposal $n$ from $L$ \& $M$}
        }
        \tcp{Now $K$ proposals for each $q$}
        \tcp{$K \le N$ after removals}
        $M^{K \times H \times W}_{q} \leftarrow$ \upsc{$M_q$}\;
        $S^{H \times W}_q \leftarrow \argmax_K (\soft_K M_q)$\;
        $I^{K \times C}_q \leftarrow l_k \,\, \forall \,\, k \in K$\;

}
$S^{Q \times H \times W} \leftarrow  \concat{S_q \,\, \forall \,\, q \in Q}$\;
$I^{Q \times K \times C} \leftarrow  \concat{I_q \,\, \forall \,\, q \in Q}$\;
\KwRet{S, I}
\end{algorithm}

\section{Implementation details and reasoning}
\label{app:sec:extra_implementation_details}
\subsection{General}
\label{app:sec:general_implementation}
We chose to use the DETR \cite{carion_end--end_2020} model as it achieves respectable performance while maintaining a simpler implementation compared to other approaches. The ResNet50 backbone was chosen as it requires fewer computational resources than other backbones (such as ResNet101) in exchange for a small drop in performance. As our focus is on uncertainty estimation rather than state-of-the-art performance, this tradeoff is acceptable. For the COCO dataset, a pre-trained model (unmodified) from \cite{carion_end--end_2020} is used. For the KITTI-STEP and VIPER datasets, which we cast as a standard panoptic segmentation task, no pre-trained model exists, so we train a model for 100 epochs with a batch size of 1.  Note that the VIPER dataset relies on a game engine to generate simulated data while the others collect real-world images. All training and evaluation for all datasets are reported on their respective dataset-provided training and validation splits (for COCO, the 2017 variant is used). Where ``corrupted'' variants of a dataset are used in some experiments, it only applies at inference time; all models are training solely on ``clean'' data. Where applicable, we convert the datasets from their original format into one matching the COCO panoptic format. Other training parameters are set following \cite{carion_end--end_2020}, including the timing of a drop in the learning rate which occurs $\frac{2}{3}$ of the way through training. Transfer learning is used: pre-trained weights provided by \cite{carion_end--end_2020} for the COCO dataset are used to initialize the model during training. One key difference, however, is that we train the models in panoptic segmentation mode only, instead of first training a bounding box model followed by a segmentation one. We found this to offer both better performance on our chosen PQ metric and require less time to train. For experiments other than the baselines, we enable dropout at test time using default value of $p = 0.1$ from \cite{carion_end--end_2020} in the encoder and decoder of the Transformer portion of the model \textit{i.e.} all layers where dropout is used in the model.

\subsection{Baseline}
\label{app:subsec:baseline_impl_details}
The baseline we compare against in all experiments involves executing the model following parameters and code from \cite{carion_end--end_2020}. On other datasets, our trained models are used. As MC Dropout is disabled at inference time, baseline results are deterministic. The baseline represents what can be achieved without the use of any samples from an ensemble, keeping all confounding variables to a minimum.

The DETR network at the most basic level generates a substantial number of false positives, so some degree of pruning is required.  The baseline accomplishes this by requiring a softmax score of at least $85\%$ and a minimum pixel count of $4$ to define a segment from a proposal. Any pixels that do not meet these thresholds are assigned to the next best segment, with the result that objects in an image where the network is uncertain (\textit{e.g.} only $75\%$ confidence) can be subsumed into another nearby region, such as \textit{road} or \textit{sky}. This brute-force approach does help the baseline when it comes to the PQ metric, but is completely unsuitable when a user wishes to know when the network is uncertain about a prediction it made. We refer to this process as \textit{pruning}, and treat this as the standard baseline. In some experiments in the paper, we use or include results from the baseline with this pruning disabled so as to provide a more appropriate comparison to our approach. Examples of this can be seen in Figures 3 and 4 in the paper.

As an additional baseline capable of generating a prediction from multiple ensemble samples, we implement a particular variant of the Hungarian method \cite{kuhn_hungarian_1955} known as the Jonker–Volgenant algorithm \cite{jonker_shortest_1987}. Our implementation is as straightforward an implementation as is possible adapting from other domains where it is used \cite{weber_step_2021}. As with our approach, it is capable of handling multiple ensemble samples and generating a final scene segmentation result. Each individual sample is processed following the baseline approach, resulting in each sample consisting of a variable number of proposals (detected objects). Each proposal has a logit and binary mask associated with it, as well as a softmax score. Initializing with a random sample, all other samples are matched against it following Jonker–Volgenant with the matching cost determined by the Intersection-over-Union of the binary masks. Upon a successful match, the binary mask area is grown by that of the matched proposal. Upon matching all of the samples as best as possible (some samples may remain unmatched due to the algorithm being designed for the classic assignment problem), a new logit mask is associated with each proposal via the mean across samples. Then, following the baseline approach, the softmax and argmax operators are applied to generate the association between pixels and detected objects.

\subsection{Basic Performance}
\label{app:subsec:basic_performance_impl_details}
In the basic performance experiment shown in Section 4.2 and Table 1 in the paper, the per-pixel uncertainty estimates are ignored. Instead, we optimize both our approach and the Hungarian algorithm such that they are presented in the best light possible respectively. As the baseline applies various thresholds to remove false positives, we also apply a basic level of pruning to our approach. Specifically, we apply a threshold on the the probability of each pixel of $0.4$ and the minimum number of pixels required to define a segment as $4$. Any pixels in the image that do not meet this criteria are left with no class or instance ID assigned - \textit{i.e.} we refuse to provide a prediction for those pixels. Ideally, as we do in later experiments, per-pixel uncertainty estimates are used to remove such false positives and other erroneous predictions.  We also apply an Intersection-over-Union threshold of $0.6$ and a minimum merge count of $80\%$ of Monte Carlo (MC) samples (\textit{e.g.} for 15 samples, 12 merges) for the instance segmentation part of our approach. Lastly, no thresholds are used in the Hungarian approach, as we found that applying any of the applicable aforementioned thresholds harmed performance. We evaluate two different values for the number of Monte Carlo samples: 5 and 15. Note that there is no consensus as to how many samples are required for any given problem, but both the original work on MC Dropout \cite{gal_dropout_2016} and much more recent work comparing against it \cite{mukhoti_deep_2021} use between 5 and 10 samples, although some use significantly more \cite{alex_kendall_bayesian_2017}. All runs using MC samples are executed 10 times, and we present the mean and standard deviation of these runs. Baseline runs are deterministic and thus presented as-is. 

\subsection{Application of uncertainty estimates}
\label{app:subsec:app_uncert_impl_details}
In this experiment, shown in Section 4.3 and Figure 3 in the paper, thresholds on probabilities, number of mergers or the minimum number of pixels are not used and only the per-pixel uncertainty estimates are used to determine when to remove pixels from evaluation. We sweep through a threshold on these numerical uncertainty estimates at each pixel, and remove the most uncertain pixels (treat as no ground truth existing) which simulates the referral of uncertain regions to a human reviewer. Thus, for different threshold values, we have varying levels of true and false positives as well as false negatives, following standard evaluation procedures \cite{kirillov_panoptic_2019}. Calculating and plotting the False Discovery Rate and True Positive Rate from this data gives us the ROC-like curves. With respect to parameters not explicitly denoted in the plot, the only one of relevance (thresholds are disabled) is the IoU threshold, which we set at $0.2$.

Note that Figure 4 is generated using the same data as Figure 3, except for either algorithm using Monte Carlo samples only a single randomly selected run is shown due to computational constraints on the amount of memory required to process all pixels from an entire dataset.

\section{Additional Results and Discussion}
\label{app:sec:additional_results}

\subsection{Hungarian Performance}
\label{app:sec:hungarian_performance}
In Table 1, we show that the Hungarian method performs worse than our approach and worse than the baseline. We believe this poor performance to be caused primarily by the Hungarian algorithm’s design of solving the classic assignment problem. As we detail in \cref{app:subsec:baseline_impl_details}, we use a popular variant of the Hungarian algorithm named the Jonker-Volgenant algorithm, which is capable of solving a more generalized version of the assignment problem with a rectangular cost matrix. This is the case applicable to our experiments as the number of detected proposals (segments) can change between samples. However, this implementation forces the least number of assignments possible, and as such some proposals in a given sample may not be matched to a corresponding proposal in another sample. We believe this least assignment requirement to be the key restriction that hinders the performance of the Hungarian algorithm. A secondary restriction of the Hungarian method is that there can only be one maximal match between proposals across samples. Our approach by contrast can match one proposal in a sample to more than one proposal in another sample if the appropriate criteria are met. This satisfies the scenario where the network may erroneously identify a single true object as multiple objects in another sample; we note this to be most common with vehicles in which individual parts can sometimes be detected as multiple independent vehicles.

\subsection{Basic Performance}
\label{app:sec:basic_performance_extra}

\begin{sidewaystable}
    \centering
    \begin{tabular}{lll|lll|lll|lll}
        \toprule
        \multicolumn{12}{c}{Segmentation Results - Clean Data}\\
        \midrule
        & App. & Samp. & PQ$_{\text{all}}\uparrow$ & SQ$_{\text{all}}\uparrow$ & RQ$_{\text{all}}\uparrow$ & PQ$_{\text{things}}\uparrow$ & SQ$_{\text{things}}\uparrow$ & RQ$_{\text{things}}\uparrow$ & PQ$_{\text{stuff}}\uparrow$ & SQ$_{\text{stuff}}\uparrow$ & RQ$_{\text{stuff}}\uparrow$\\
        \midrule
        \multirow{5}{*}{\rotatebox{90}{COCO}} & Baseline & N/A & 43.4 & 79.3 & 53.8 & 48.2 & 79.8 & 59.5 & 36.3 & \textbf{78.5} & 45.3\\
        & Hungarian & 5 & 39.5±0.1 & 78.8±0.1 & 49.0±0.1 & 45.3±0.2 & 79.7±0.1 & 55.9±0.2 & 30.7±0.1 & 77.4±0.2 & 38.7±0.1\\
        & Hungarian & 15 & 40.7±0.1 & 79.1±0.1 & 50.4±0.1 & 46.3±0.1 & 79.9±0.1 & 57.0±0.2 & 32.2±0.1 & 77.8±0.1 & 40.4±0.1\\
        & Ours & 5 & 44.1±0.1 & 79.5±0.1 & \textbf{54.5±0.1} & 48.9±0.1 & 80.5±0.1 & \textbf{59.9±0.1} & 36.9±0.1 & 77.9±0.1 & 46.3±0.2\\
        & Ours & 15 & \textbf{44.4±0.1} & \textbf{79.9±0.1} & \textbf{54.5±0.1} & \textbf{49.1±0.2} & \textbf{81.0±0.1} & 59.7±0.2 & \textbf{37.2±0.1} & 78.2±0.1 & \textbf{46.6±0.1}\\
        \midrule
        \multirow{5}{*}{\rotatebox{90}{KITTI}} & Baseline & N/A & 39.9 & \textbf{77.5} & 50.2 & 57.2 & 84.2 & 67.5 & 37.9 & 76.7 & 48.1\\
        & Hungarian & 5 & 36.1±0.2 & 76.5±0.2 & 45.4±0.2 & 53.0±0.1 & 83.9±0.0 & 62.8±0.2 & 34.1±0.2 & 75.7±0.2 & 43.4±0.2\\
        & Hungarian & 15 & 37.6±0.1 & 76.9±0.1 & 47.2±0.1 & 54.5±0.1 & 84.1±0.0 & 64.4±0.2 & 35.6±0.1 & 76.0±0.1 & 45.2±0.2\\
        & Ours & 5 & \textbf{41.8±0.1} & \textbf{76.9±0.1} & \textbf{52.7±0.2} & 57.7±0.1 & 84.8±0.1 & 67.6±0.2 & \textbf{39.9±0.1} & 75.9±0.1 & \textbf{50.9±0.2}\\
        & Ours & 15 & \textbf{41.9±0.1} & 77.0±0.1 & \textbf{52.7±0.1} & \textbf{58.1±0.1} & \textbf{85.2±0.0} & \textbf{67.9±0.1} & \textbf{39.9±0.1} & \textbf{76.0±0.1} & \textbf{50.9±0.1}\\
        \midrule
        \multirow{5}{*}{\rotatebox{90}{VIPER}}
        & Baseline & N/A & 31.1 & 73.9 & \textbf{39.7} & 22.6 & 74.4 & 29.7 & \textbf{37.7} & 73.6 & \textbf{47.4}\\
        & Hungarian & 5 & 27.0±0.0 & 73.1±0.1 & 34.6±0.0 & 21.0±0.1 & 73.9±0.1 & 27.8±0.1 & 31.6±0.0 & 72.5±0.1 & 39.8±0.1\\
        & Hungarian & 15 & 28.4±0.0 & 73.6±0.1 & 36.3±0.0 & 21.9±0.0 & 74.2±0.1 & 28.9±0.1 & 33.4±0.0 & 73.1±0.1 & 42.0±0.0\\
        & Ours & 5 & 31.1±0.0 & 74.2±0.1 & 39.5±0.0 & 22.9±0.0 & 75.0±0.1 & \textbf{29.9±0.1} & 37.4±0.0 & \textbf{73.7±0.1} & 46.9±0.0\\
        & Ours & 15 & \textbf{31.3±0.0} & \textbf{74.6±0.0} & 39.6±0.0 & \textbf{23.0±0.0} & \textbf{75.7±0.0} & 29.7±0.1 & 37.6±0.0 & \textbf{73.7±0.0} & 47.2±0.0\\
        \bottomrule
    \end{tabular}
    \caption{Basic performance. Results are presented as $\mu\,$±$\,\sigma$ over 10 runs when stochastic.}
    \label{app:tab:eval_table_big_clean}
\end{sidewaystable}

\begin{sidewaystable}
    \centering
    \begin{tabular}{lll|lll|lll|lll}
        \toprule
        \multicolumn{12}{c}{Segmentation Results - Gaussian and Shot Noise Severity 1}\\
        \midrule
        & App. & Samp. & PQ$_{\text{all}}\uparrow$ & SQ$_{\text{all}}\uparrow$ & RQ$_{\text{all}}\uparrow$ & PQ$_{\text{things}}\uparrow$ & SQ$_{\text{things}}\uparrow$ & RQ$_{\text{things}}\uparrow$ & PQ$_{\text{stuff}}\uparrow$ & SQ$_{\text{stuff}}\uparrow$ & RQ$_{\text{stuff}}\uparrow$\\
        \midrule
        \multirow{5}{*}{\rotatebox{90}{COCO}} & Baseline & N/A & 29.0 & 75.4 & 36.8 & 33.6 & 76.4 & 42.8 & 21.9 & 73.8 & 27.8\\
        & Hungarian & 5 & 28.1±0.0 & 76.2±0.2 & 35.7±0.1 & 33.2±0.1 & 76.7±0.4 & 42.1±0.1 & 20.3±0.1 & 75.3±0.2 & 26.1±0.2\\
        & Hungarian & 15 & 28.9±0.1 & 76.3±0.2 & 36.7±0.1 & 34.0±0.1 & 76.6±0.0 & 43.0±0.1 & 21.2±0.1 & 75.7±0.4 & 27.1±0.2\\
        & Ours & 5 & \textbf{30.6±0.1} & 76.8±0.2 & \textbf{38.7±0.1} & \textbf{35.2±0.1} & 77.4±0.3 & \textbf{44.2±0.1} & 23.7±0.1 & 75.8±0.1 & 30.3±0.1\\
        & Ours & 15 & 30.5±0.1 & \textbf{77.1±0.1} & 38.4±0.1 & 35.0±0.1 & \textbf{77.8±0.1} & 43.6±0.1 & \textbf{23.9±0.0} & \textbf{76.0±0.1} & \textbf{30.5±0.1}\\
        \midrule
        \multirow{5}{*}{\rotatebox{90}{KITTI}} & Baseline & N/A & 19.3 & 54.9 & 26.0 & 42.0 & 79.1 & 52.5 & 16.6 & 52.1 & 22.9\\
        & Hungarian & 5 & 17.1±0.1 & \textbf{58.3±1.2} & 23.1±0.1 & 40.4±0.2 & 78.7±0.1 & 50.7±0.21 & 14.3±0.1 & \textbf{55.9±1.4} & 19.9±0.1\\
        & Hungarian & 15 & 17.8±0.1 & \textbf{58.1±0.2} & 24.1±0.1 & 41.1±0.1 & 78.8±0.1 & 51.5±0.2 & 15.1±0.1 & \textbf{55.7±0.2} & 20.9±0.1\\
        & Ours & 5 & 20.0±0.0 & 57.4±1.1 & \textbf{27.1±0.0} & \textbf{42.5±0.1} & 79.2±0.1 & \textbf{53.1±0.2} & 17.4±0.0 & 54.8±1.3 & 24.0±0.0\\
        & Ours & 15 & \textbf{20.1±0.0} & 57.7±0.9 & \textbf{27.1±0.0} & \textbf{42.5±0.1} & \textbf{79.4±0.1} & 52.9±0.1 & \textbf{17.5±0.0} & 55.1±1.0 & 24.1±0.1\\
        \midrule
        \multirow{5}{*}{\rotatebox{90}{VIPER}}
        & Baseline & N/A & \textbf{10.3} & 66.8 & \textbf{14.5} & 8.9 & 74.1 & 12.1 & \textbf{11.4} & 61.1 & \textbf{16.4}\\
        & Hungarian & 5 & 8.4±0.0 & 68.8±0.2 & 12.1±0.0 & 9.5±0.0 & 73.5±0.1 & 13.0±0.1 & 7.6±0.0 & 65.2±0.3 & 11.4±0.0\\
        & Hungarian & 15 & 9.2±0.0 & 69.2±0.2 & 13.1±0.0 & \textbf{9.8±0.0} & 73.9±0.2 & \textbf{13.3±0.1} & 8.7±0.0 & 65.5±0.2 & 12.9±0.0\\
        & Ours & 5 & 10.2±0.0 & 69.5±0.1 & 14.3±0.0 & 9.4±0.0 & 74.4±0.2 & 12.7±0.0 & 10.8±0.0 & \textbf{65.8±0.2} & 15.6±0.0\\
        & Ours & 15 & 10.1±0.0 & \textbf{69.7±0.2} & 14.2±0.0 & 9.0±0.0 & \textbf{75.0±0.1} & 12.1±0.0 & 11.0±0.0 & 65.6±0.2 & 15.9±0.1\\
        \bottomrule
    \end{tabular}
    \caption{Basic performance. Results are presented as $\mu\,$±$\,\sigma$ over 10 runs when stochastic.}
    \label{app:tab:eval_table_big_corrupt}
\end{sidewaystable}

In \cref{app:tab:eval_table_big_clean,app:tab:eval_table_big_corrupt}, we expand Table 1 in the paper with a further breakdown of the metric into the \textit{Stuff} and \textit{Things} sub-categories for both clean and corrupt data. As we previously discussed, \textit{things} includes objects that can be distinguished from one another and are labelled as such in the dataset, such as cars. \textit{Stuff} includes objects that can not be identified as individual objects, or has been deemed to difficult to do so by the dataset annotators, with categories such as \textit{building} or \textit{sky} being common examples. 

In \cref{app:tab:eval_table_big_clean,app:tab:eval_table_big_corrupt}, broadly speaking the conclusions that can be drawn are the same as presented in the paper. Beyond that, in \cref{app:tab:eval_table_big_clean} we see that performance differences between approaches are broadly replicated across both the things and stuff sub-categories, with the exception on the VIPER dataset where our method does better overall but fails to match the baseline or improve on it by a slim margin. In \cref{app:tab:eval_table_big_corrupt}, the same applies: performance increases or decreases in both \textit{stuff} and \textit{things} categories generally track with the VIPER dataset being an exception. We also see that the added noise decimates performance pretty much across the board, and results in some notable exceptions such as the Hungarian method doing the best in the PQ$_{\text{things}}$ category, but performance is so poor (especially compared to clean data) that it is not appropriate to draw any conclusions from such exceptions.

\subsection{Qualitative Results}
\label{app:sec:qual_extended}
\begin{figure*}
    \centering
\begin{subfigure}{0.48\textwidth}
    \includegraphics[width=\textwidth]{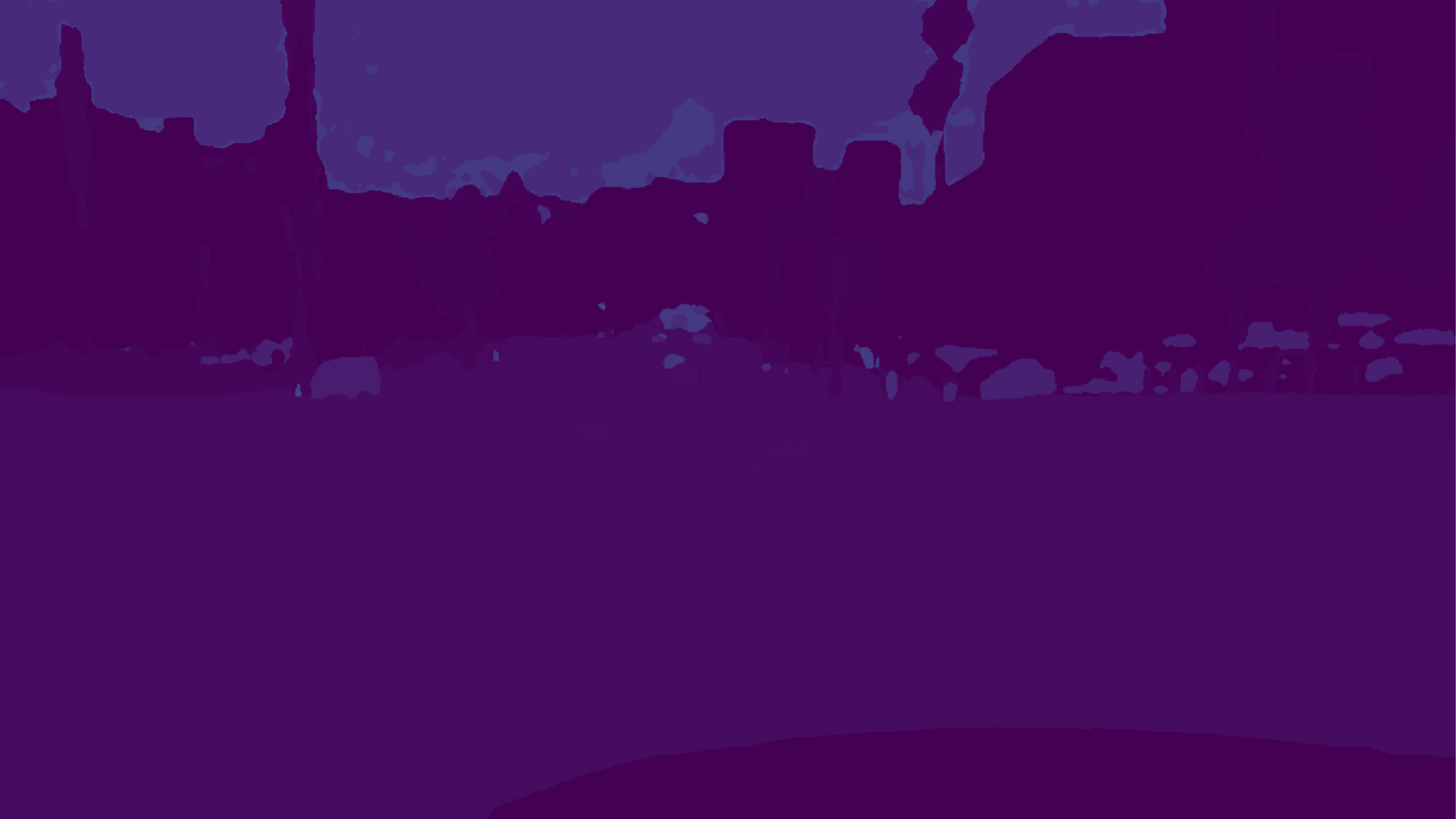}
\end{subfigure}
\begin{subfigure}{0.48\textwidth}
    \includegraphics[width=\textwidth]{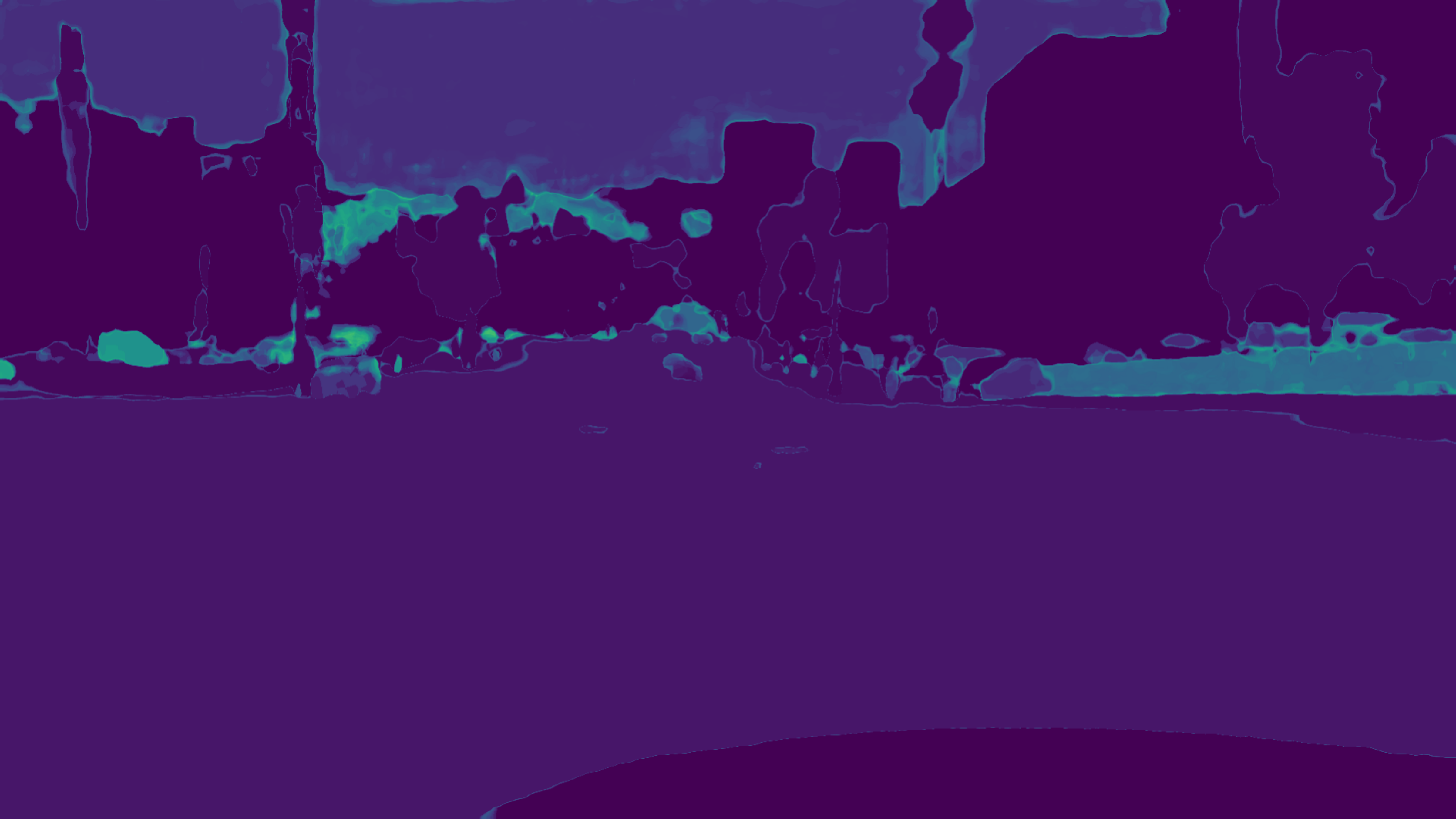}
\end{subfigure}
    \caption{Side-by-side comparisons of predictive entropy uncertainty heatmaps for the baseline (left) and our approach with 15 MC samples (right) for the bottom-most example shown in Fig. 4 in the paper. Lighter shades indicate greater uncertainty.}
    \label{app:fig:qual_entropy_comparison_VIPER}
\end{figure*}

In \cref{app:fig:qual_entropy_comparison_VIPER}, we expand on Figure 6 from the paper by including an example from the VIPER dataset to demonstrate that the differences are not solely restricted to one dataset.

\subsection{Execution time}
\label{app:subsec:execution_time}
\begin{figure}
    \centering
    \includegraphics[width=\linewidth]{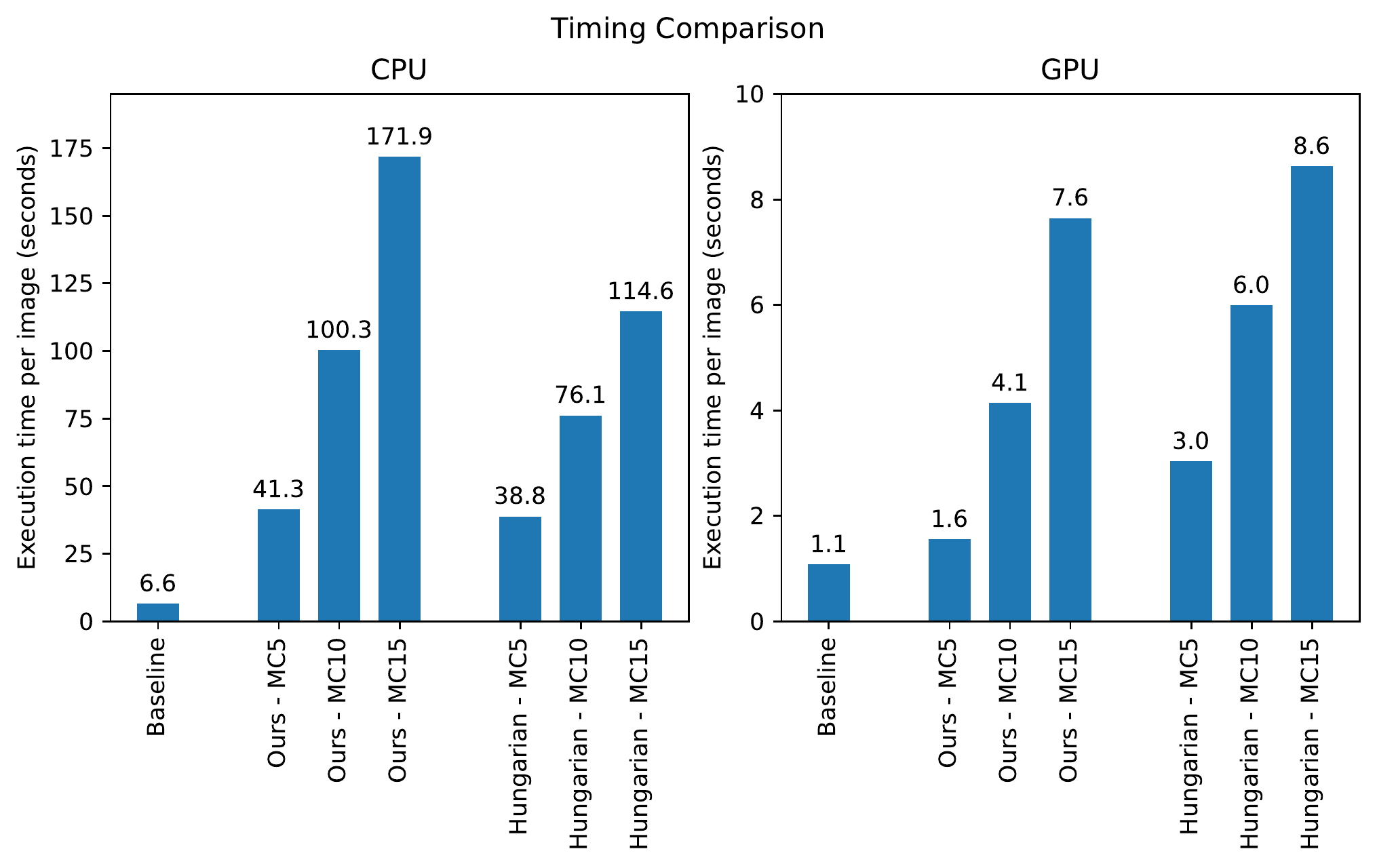}
    \caption{Execution time comparison of the baseline, our approach and the Hungarian algorithm, with varying numbers of Monte Carlo samples, presented as the mean over 5 runs. The variance between runs is visually negligible and thus not shown.}
    \label{app:fig:runtime_comparison}
\end{figure}

In \cref{app:fig:runtime_comparison} we compare the baseline, our approach and the Hungarian algorithm in terms of the runtime in seconds per image, for both CPU and GPU implementations for varying numbers of Monte Carlo samples. Note that we include time to load data into the GPU and relevant post-processing steps and exclude loading the model from disk and the calculation of evaluation metrics. Results are calculated using a selected set of 100 images from the COCO dataset, and executed on a Intel Xeon E5-1620 v3 with 94GB of RAM and a Nvidia GeForce RTX 2080 Ti. In general, the code is research-level rather than production-level, and as such the timing shown here cannot be broadly generalized without a thorough understanding of the software frameworks and relevant hardware available. Further to that, one important caveat is that our implementation of the Hungarian algorithm is CPU only, whereas our approach can run fully on the GPU.

On the CPU side we can see that our approach is slightly more costly than the Hungarian method, and more so as the number of samples grows. The Hungarian method does however pay for this advantage in terms of metric performance as we show in Table 1 in the paper. In general, however, any approach here other than the baseline is not very practical to use, as waiting more than half a minute per image is excessive for a number of use cases, such as autonomous driving.

On the GPU side, things are a little more interesting. First, we can see that the relationship between the number of Monte Carlo samples and runtime is linear but only beyond a certain number of samples. We believe that bottlenecks and optimizations in certain areas such as the execution of the DETR model contribute to the fact that using 5 samples is not much more costly than the single sample used by the baseline, for either our approach or the Hungarian algorithm. This means that using at least a few samples is at least feasible in production environment with timing requirements, given some optimization work. Second, the Hungarian algorithm in particular performs worse than our approach, but it is still not five times as costly as the baseline when using 5 samples. Any firm conclusions however comparing our approach and the Hungarian algorithm cannot be made due to the aforementioned implementation limitations.

Overall, we can say that the practical cost of using ensembles and our approach is not as high as assumed by some (\textit{e.g.} \cite{sirohi_uncertainty-aware_2023}) but not as low as assumed by others (\textit{e.g.} \cite{gal_dropout_2016}), and rather somewhere in-between. These findings are of course contingent on the specific implementation and hardware available.

\end{document}